\documentclass[11pt]{article}
\usepackage{macros}

\title{
\texttt{VDSAgents}: A PCS-Guided Multi-Agent System for Veridical Data Science Automation
}

\author{
Yunxuan Jiang \thanks{School of Management, Xi’an Jiaotong University; \texttt{fengjianliu@stu.xjtu.edu.cn}. } \\
\and Silan Hu \thanks{School of Computing, National University of Singapore; \texttt{silan.hu@u.nus.edu}. \\~~Yunxuan Jiang and Silan Hu have equally contributed to this work.} \\
\and Xiaoning Wang \thanks{School of Data Science and Media Intelligence, Communication University of China; \texttt{sdwangxiaoning@cuc.edu.cn}.}\\ 
\and Yuanyuan Zhang \thanks{Beijing Baixingkefu Network Technology Co., Ltd.;  \texttt{zhang.huanzhiyuan@gmail.com}.}\\
\and Xiangyu Chang \thanks{School of Management, Xi’an Jiaotong University; \texttt{xiangyuchang@xjtu.edu.cn}. 
} 
}
\begin{document}

\maketitle

\begin{abstract}%
Large language models (LLMs) become increasingly integrated into data science workflows for automated system design. 
However, these LLM-driven data science systems rely solely on the internal reasoning of LLMs, lacking guidance from scientific and theoretical principles. 
This limits their trustworthiness and robustness, especially when dealing with noisy and complex real-world datasets.
This paper provides \texttt{VDSAgents}\footnote{VDSAgents projection is publicly available at \url{https://github.com/fengzer/VDSAgents}.}, a multi-agent system grounded in the Predictability-Computability-Stability (PCS) principles~\citep{Yu2020VDS} proposed in the Veridical Data Science (VDS)~\citep{vdsbook}.
Guided by PCS principles, the system implements a modular workflow for data cleaning, feature engineering, modeling, and evaluation.
Each phase is handled by an elegant agent, incorporating perturbation analysis, unit testing, and model validation to ensure both functionality and scientific auditability.
We evaluate \texttt{VDSAgents} on nine datasets with diverse characteristics, comparing it with state-of-the-art end-to-end data science systems, such as 
\texttt{AutoKaggle} and \texttt{DataInterpreter}, using DeepSeek-V3 and GPT-4o as backends.
\texttt{VDSAgents} consistently outperforms the results of \texttt{AutoKaggle} and \texttt{DataInterpreter}, which validates the feasibility of embedding PCS principles into LLM-driven data science automation.
\end{abstract}
\section{Introduction}

\begin{figure}[!ht]
\centering
\includegraphics[width=0.85\linewidth]{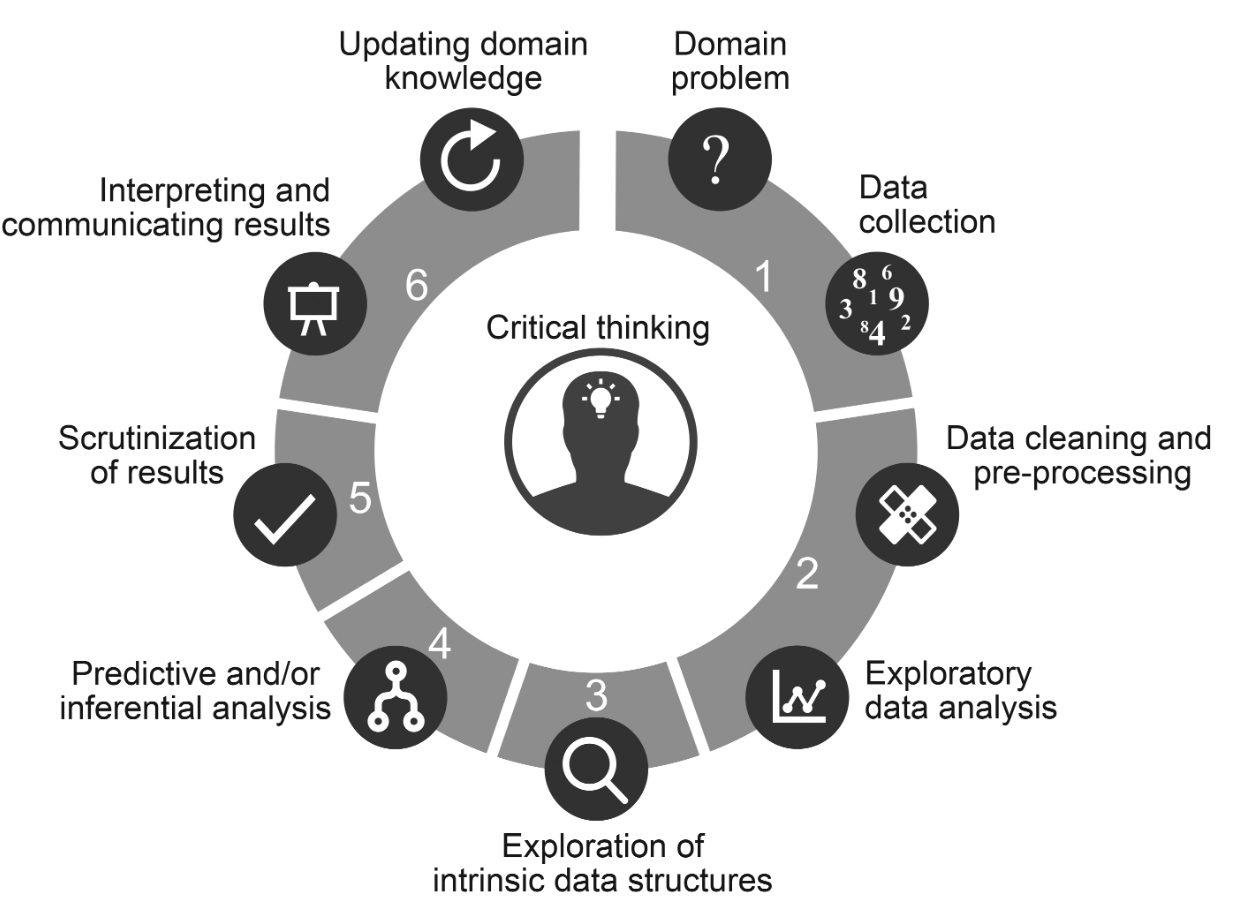}
\caption{Illustration of the DSLC. It includes six stages: (1) identifying and formulating domain problems and collecting data; (2) data cleaning, pre-processing, and early exploration; (3) optional structural analysis and data mining; (4) optional modeling and statistical inference; (5) evaluation and validation of results; and (6) interpretation, communication, and domain knowledge update.}
\label{fig_ds}
\end{figure}

Data science has emerged as a multidisciplinary field that integrates statistics, computer science, mathematics, and domain knowledge to extract meaningful insights and guide decision-making from complex data~\citep{Yu2020VDS}.
Its scope spans the entire data science lifecycle (DSLC), from collection and pre-processing to modeling, validation, and knowledge refinement, and plays a vital role in decision-making in scientific, industrial, and policy domains~\citep{Cao2017,provost2013science}.
A typical DSLC is illustrated in Figure~\ref{fig_ds} proposed by~\citet{vdsbook}, which outlines the key stages of data science-based research.
As data continues to grow in volume, complexity, and heterogeneity, standard data science approaches are increasingly insufficient to meet the demands for trustworthiness and robustness. 
This has fueled the development of automated and principled DSLC. 

Recent advances in large language models (LLMs), especially frontier models like GPT-4~\citep{achiam2023gpt} and DeepSeek-V3~\citep{deepseek-llm}, have significantly reshaped the landscape of data science automation. 
Using prompt engineering, tool integration, and code generation capabilities, LLMs have been incorporated into systems that perform various stages of the data science pipeline~\citep{Li2024AutoKaggle,Sun2024LAMBDA,Hong2024DataInterpreter}. 
A new paradigm has emerged, the agent-based approach, in which LLMs are organized into structured, role-based entities capable of simulating data scientists in end-to-end workflows. 
As~\citet{Tu2024LLMDataScience} observe, LLMs are increasingly positioned as strategic collaborators, shifting practitioners from manual operations to high-level planning and oversight.

Despite the promise of this approach, current agent-based data science systems face several persistent challenges. 
Existing frameworks like \texttt{AutoKaggle}~\citep{Li2024AutoKaggle}, \texttt{LAMBDA}~\citep{Sun2024LAMBDA}, and \texttt{DataInterpreter}~\citep{Hong2024DataInterpreter} depend primarily on the intrinsic reasoning ability of the LLM to plan and execute multistep tasks. However, this autonomy often results in brittle execution paths, low reproducibility, and limited robustness, especially in the presence of noisy, missing, or structurally inconsistent data.
These systems typically lack principled mechanisms to guide workflow design, evaluate stability, or explore alternative analytic decisions. 
Consequently, while they are capable of producing seemingly valid pipelines, they often fail to ensure trustworthiness or consistency across datasets and contexts.

The Veridical Data Science (VDS) framework~\citep{Yu2020VDS} provides a theory-based foundation for addressing these issues.
Based on Predictability-Computability-Stability (PCS) principles, VDS advocates data science as a process of critical and transparent reasoning rather than a mere algorithm execution.
It emphasizes the importance of systematically examining model choices, testing stability via perturbation, and ensuring analytic reproducibility. 
In this work, we propose integrating VDS into the architecture of LLM-agent systems by utilizing it as a structured external planning framework.
This work leverages PCS principles and proposes a novel agent-based data science framework, which is naturally referred to as \texttt{VDSAgents}.
Rather than relying solely on LLMs to autonomously generate task plans, we provide a predefined multistage skeleton grounded in DSLC (see Figure \ref{fig_ds}). 
This structure divides the pipeline into phases, including problem formulation, data cleaning and exploration, feature engineering and modeling, and result evaluation. 
Each stage is managed by a dedicated agent, and a central \texttt{PCS-Agent} operates across all phases to assess and improve the predictability, computability, and stability of the overall workflow.
The \texttt{PCS-Agent} offers theoretical feedback at all levels of the workflow—questioning data credibility, suggesting alternative problem framings, and enforcing reproducibility checks. 
This design enables the system to systematically explore diverse analytic paths, explicitly model uncertainty, and ensure more reliable results using PCS principles~\citep{vdsbook}.

The main contributions of this paper are summarized as follows:

\begin{itemize}
  \item \textbf{\texttt{VDSAgents} framework:} We propose the first multi-agent system that systematically integrates the DSLC into LLM-based architectures, guided by the PCS principles.
  The framework designs a dedicated \texttt{PCS-Agent} to guide all other agents in the DSLC.

  \item \textbf{Scientific tool integration:} A modular tool set is developed to support code execution, unit testing, fault diagnosis, and image-to-text transformation, enhancing robustness and flexibility.

  \item \textbf{Paradigm advancement:} This work proposes a new paradigm of automation for trustworthy AI-assisted data science, bridging VDS and LLM agent methodologies.
\end{itemize}

We validate the proposed framework through systematic experiments on real-world datasets. 
Our results demonstrate that \texttt{VDSAgents} achieves superior robustness and predictive performance compared to representative LLM-driven systems.
\section{Related Work}

\subsection{Large Language Models for Data Science}

LLMs have demonstrated powerful capabilities in natural language understanding, reasoning, and code generation~\citep{achiam2023gpt,deepseek-llm}.
These advances have led to their growing use in data science, where LLMs can assist with tasks such as data cleaning, exploratory data analysis (EDA), feature engineering, modeling, and automated report writing~\citep{Li2024AutoKaggle,Hong2024DataInterpreter,Sun2024LAMBDA,aide2025}. Their ability to follow natural language instructions enables users to perform complex analyses with minimal coding.

As LLMs become more embedded in data science workflows, the focus shifts from manual execution to oversight and validation~\citep{Tu2024LLMDataScience}.
To support this transition, there is a pressing need to introduce external theoretical frameworks that ensure the transparency, stability, and reproducibility of LLM-driven data analysis.

\subsection{VDS and PCS Principle}
VDS is a principled framework proposed by~\citet{Yu2020VDS} to promote robust, reliable, and reproducible data science. 
It centers around the PCS principles: predictability, computability, and stability, each addressing critical aspects of reliable data analysis~\citep{Yu2013Stability}. Predictability ensures that models generalize to new data; Computability emphasizes practical feasibility; and Stability tests the sensitivity of results to data and decision perturbations.

Despite its growing influence on human-led workflows, the VDS framework has not yet been integrated into autonomous LLM-driven agent systems.
We argue that PCS principles provide valuable guidance for managing uncertainty and enhancing reproducibility in LLM-driven pipelines.
Our work makes the first attempt to systematically apply PCS principles to guide agent behavior across the entire DSLC.

\subsection{Multi-Agent Systems and Task Planning}

Many LLM-based systems rely on internal planning methods such as ReAct~\citep{yao2022react} and Tree-of-Thoughts (ToT)~\citep{yao2023tree} to structure reasoning and execution.
However, these approaches often suffer from instability and lack of reproducibility in DSLC, where task dependencies are complex and results must be tightly controlled.

To mitigate these issues, recent frameworks have adopted multi-agent designs with explicit task decomposition.
\texttt{AutoKaggle}~\citep{Li2024AutoKaggle} structures the pipeline into dedicated agents for data pre-processing, modeling, and evaluation, improving modularity and execution traceability.
\texttt{LAMBDA}~\citep{Sun2024LAMBDA} similarly defines role-specific agents to coordinate modeling tasks.
\texttt{DataInterpreter}~\citep{Hong2024DataInterpreter} further enhances coordination with a hierarchical task graph that supports dynamic planning and revision at all stages.

These systems demonstrate that combining structured planning with agent-based collaboration can improve interpretability and robustness.
Our framework extends this principle by integrating an external \texttt{PCS-Agent} as a critical thinker to guide multi-agent execution across the full DSLC.

\subsection{Tool Integration and Execution Reliability}

Recent research shows that the integration of external mechanisms, such as unit testing, execution feedback, and self-refinement, can significantly improve the reliability of LLM in complex tasks~\citep{Madaan2023SelfRefine}.
For example, \texttt{AutoKaggle}~\citep{Li2024AutoKaggle} incorporates an executor with error capture capabilities that detects runtime failures and automatically triggers correction procedures, thus improving both task completion rates and execution stability.
In addition, other studies emphasize the use of tool-enhanced pipelines for verification and debugging~\citep{wang2023selfdebug,zhou2023whiteboxtest}.
Beyond using predefined tools, recent approaches also allow LLMs to dynamically create, manage, or adapt tools for specific tasks~\citep{cai2023toolmaker,schick2023toolformer,qian2023creator}.

Unit testing has emerged as a practical approach to validating the logic of LLM-generated code~\citep{zhou2023autoagents}.
Frameworks such as PAL~\citep{Gao2022PAL} embed intermediate programmatic reasoning and use test cases to verify the correctness of intermediate steps.
This helps ensure the internal consistency and verifiability of multi-step reasoning processes.

Our system integrates a modular and extensible toolset to support reliable execution.
This includes components for code execution, fault detection, self-debugging, OCR-based image-to-text conversion, and unit testing.
All tools are designed to be dynamically callable by agents and are decoupled from specific back-end models, enabling flexible deployment across different environments.

In summary, LLM-driven data science systems are evolving toward greater structure, moving from single-model reasoning to multi-agent collaboration and tool-assisted execution. 
Although these systems have improved efficiency and task coverage, they still struggle with the complexity of real-world data, reasoning stability, and the trustworthiness of results. 
Existing solutions, such as unit testing and workflow supervision, offer partial improvements but often lack a theoretical foundation. 
To address these challenges, we propose \texttt{VDSAgents}, a multi-agent framework guided by the PCS principles, aiming to support trustworthy, stable, and reproducible automated DSLC.
\section{Methodology}

\subsection{Overview of \texttt{VDSAgents}}

In real-world data science scenarios, challenges such as complex task dependencies, inconsistent data quality, and subjective modeling choices often compromise the trustworthiness, reproducibility, and robustness of results. 
To address these issues, \texttt{VDSAgents} decomposes the workflow into the following five dedicated agents.

\begin{itemize}
  \item \texttt{Define-Agent}: \textbf{$\mathcal{A}_\mathrm{define}$} — Formulates the problem and evaluates the data quality;
  \item \texttt{Explore-Agent}: \textbf{$\mathcal{A}_\mathrm{explore}$} — Handles data cleaning, preprocessing, and exploratory analysis;
  \item \texttt{Model-Agent}: \textbf{$\mathcal{A}_\mathrm{model}$} — Conducts feature engineering, model training, and prediction;
  \item \texttt{Evaluate-Agent}: \textbf{$\mathcal{A}_\mathrm{evaluate}$} — Assesses model performance and interprets results;
  \item \texttt{PCS-Agent}: \textbf{$\mathcal{A}_\mathrm{PCS}$} — Operates across all stages, enforcing predictability, computability, and stability through perturbation analysis and reproducibility checks.
\end{itemize}

\begin{figure}[ht]
  \centering
  \includegraphics[width=1\linewidth]{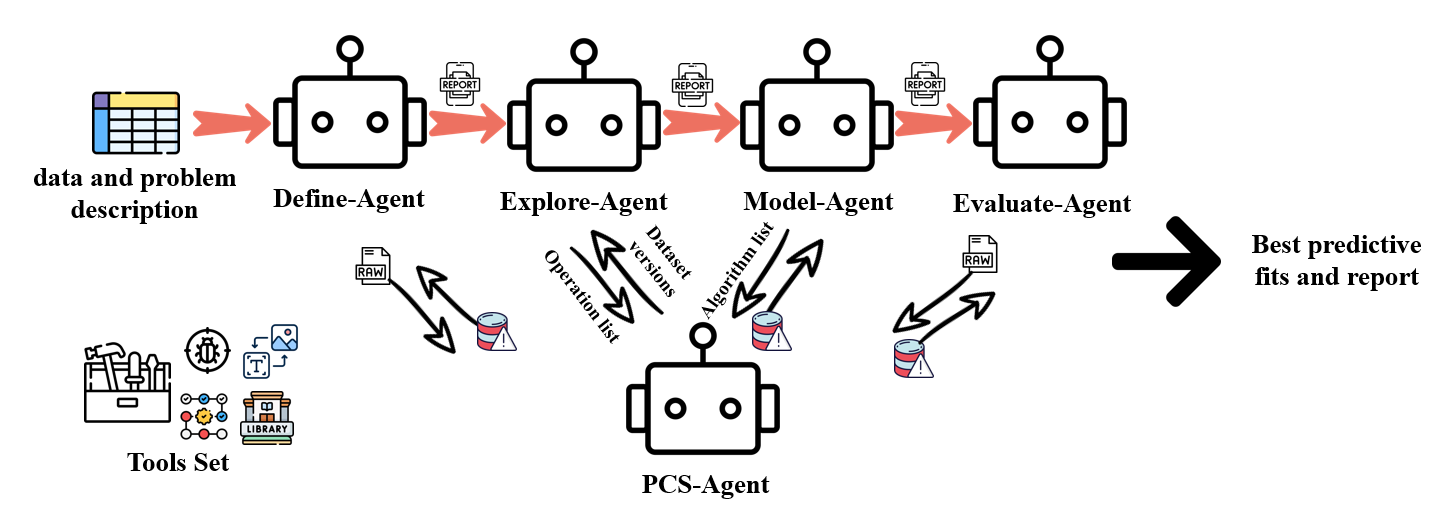}
  \caption{Workflow architecture of the \texttt{VDSAgents}. 
  Note that the \texttt{PCS-Agent} interacts with all stage-specific agents to evaluate predictability, computability, and stability.}
  \label{fig:vds_framework}
\end{figure}

Figure~\ref{fig:vds_framework} illustrates the high-level system architecture. The system has two key design components:
\begin{itemize}
    \item \textbf{PCS-Guided Workflow.}  
The data science process is divided into five sequential stages: problem definition and evaluation of data quality, data cleaning and EDA, predictive modeling, evaluation of results, and PCS-Guided perturbation and comparison. 
This structure ensures that each step is scientifically grounded and aligned with the PCS principles.
\item \textbf{Modular Multi-Agent Architecture.}  
Each agent is responsible for executing a specific phase of the workflow and operates using \textit{statements from the VDS book~\citep{vdsbook} as prompts} and shared memory.
The \texttt{PCS-Agent} continuously analyzes intermediate outputs, performs perturbation testing, and evaluates the stability and consistency of the results.
\end{itemize}

To further formalize the execution logic, we define a high-level algorithm (Algorithm~\ref{alg:vdsagent}) that governs the interactions between agents and tool usage across stages. 
Let $\Phi = \{\phi_1, \phi_2, \phi_3, \phi_4\}$ represent the different stages of problem definition and evaluation of data quality, data cleaning and EDA, predictive modeling, and evaluation of results, $s_t$ the state of the system at time $t$, $r_a$ the output of agent $a$, $T$ the set of unit tests in the current stage, $k$ the number of perturbed datasets generated for robustness analysis, $l$ the number of candidate models trained, and $\hat{Y}_{\text{test}}$ the predictions of the final model. 
Here, $\mathrm{delete}(D_{\mathrm{clean}})$ denotes the removal of both the intermediate cleaned dataset and the generated code that produced it, upon unit-test failure.

The toolset $\mathcal{T}_i$ includes task-specific utilities such as a converter code executor, debug tool, machine learning (ML) library, and image-to-text, which allows for stage-specific automation and recovery.

\begin{algorithm}[ht]
\caption{\texttt{VDSAgents} Workflow}
\label{alg:vdsagent}
\KwIn{Raw dataset $D_{\mathrm{raw}}$}
\KwOut{Structured analysis report $\mathcal{R}$ and predictions $\hat{Y}_{\mathrm{test}}$}

Initialize system state $s_0 \leftarrow$ task description and problem definition\;
Define stage sequence $\Phi = \{\phi_1, \phi_2, \phi_3, \phi_4\}$\;
Define agent set $\mathcal{A} = \{\mathcal{A}_\mathrm{define}, \mathcal{A}_\mathrm{explore}, \mathcal{A}_\mathrm{model}, \mathcal{A}_\mathrm{evaluate}, \mathcal{A}_\mathrm{PCS}\}$\;

\ForEach{stage $\phi \in \Phi$}{
    \uIf{$\phi = \phi_1$}{
        $\mathcal{T}_1 = \{\textnormal{code executor}, \textnormal{debug tool}\}$\;
        $r_1 \leftarrow \mathcal{A}_\mathrm{define}.\mathrm{execute}(D_{\mathrm{raw}}, \mathcal{T}_1)$\;
        $\mathcal{A}_\mathrm{PCS}.\mathrm{analyze}(r_1, \mathcal{T}_1)$\;
    }
    \uElseIf{$\phi = \phi_2$}{
        $\mathcal{T}_2 = \{\textnormal{code executor}, \textnormal{debug tool}, \textnormal{ML tools}, \textnormal{image-to-text}\}$\;
        $D_{\mathrm{clean}} \leftarrow \mathcal{A}_\mathrm{explore}.\mathrm{clean}(D_{\mathrm{raw}}, \mathcal{T}_2)$\;
        \While{$\neg \mathrm{unitTestsPassed}(D_{\mathrm{clean}})$}{
        $\mathcal{A}_\mathrm{explore}.\mathrm{delete}(D_{\mathrm{clean}})$}\;
        $D_{\mathrm{clean}} \leftarrow \mathcal{A}_\mathrm{explore}.\mathrm{clean}(D_{\mathrm{raw}}, \mathcal{T}_2)$\;
        
        $E \leftarrow \mathcal{A}_\mathrm{explore}.\mathrm{EDA}(D_{\mathrm{clean}}, \mathcal{T}_2)$\;
        $D_{1}, ..., D_{k} \leftarrow \mathcal{A}_\mathrm{PCS}.\mathrm{perturb}(D_{\mathrm{clean}}, k)$\;
    }
    \uElseIf{$\phi = \phi_3$}{
        $\mathcal{T}_3 = \{\textnormal{code executor}, \textnormal{debug tool}, \textnormal{ML tools}\}$\;
        $\mathcal{M}_{1:l} \leftarrow \mathcal{A}_\mathrm{model}.\mathrm{train}(D_{\mathrm{clean}}, \mathcal{T}_3)$\;
        $\mathrm{results} \leftarrow \mathcal{A}_\mathrm{PCS}.\mathrm{reproduce}(D_{1:k}, \mathcal{M}_{1:l}, \mathcal{T}_3)$\;
        $\mathcal{A}_\mathrm{PCS}.\mathrm{selectTopK}(\mathrm{results})$\;
        $\mathcal{A}_\mathrm{PCS}.\mathrm{generateReport}()$\;
    }
    \Else{
        $\mathcal{T}_4 = \{\textnormal{code executor}, \textnormal{debug tool}\}$\;
        $\hat{Y}_{\mathrm{test}} \leftarrow \mathcal{A}_\mathrm{evaluate}.\mathrm{model}(D_{\mathrm{test}}, \mathcal{M}_{\mathrm{best}}, \mathcal{T}_4)$\;
    }
}
\Return{$\mathcal{R}, \hat{Y}_{\mathrm{test}}$}\;
\end{algorithm}

\subsection{PCS-Guided Planning and Perturbation}

A key innovation of the \texttt{VDSAgents} lies in embedding PCS-Guided planning in each agent through structured prompting and shared memory.
This ensures that the agents operate not only reactively, but also in accordance with principled scientific reasoning.

Each agent receives two layers of prompt instructions:  
\begin{itemize}
  \item \textbf{System Message:} defines the agent’s role, scope of action, and associated PCS principle;
  \item \textbf{Task-Specific Message:} includes upstream outputs, stage-specific objectives, expected output formats (e.g., Python, Markdown, JSON), and relevant domain constraints.
\end{itemize}

Agents also maintain an intermediate memory state, allowing them to incorporate decisions made in earlier stages and reason within the context of the entire pipeline.
This forms a cohesive planning skeleton aligned with the DSLC.
Among all agents, the \texttt{PCS-Agent} serves as the central coordinator: it applies critical thinking grounded in the PCS principles to continuously guide and evaluate the outputs of $\mathcal{A}_\mathrm{define}$, $\mathcal{A}_\mathrm{explore}$, $\mathcal{A}_\mathrm{model}$ and $\mathcal{A}_\mathrm{evaluate}$, ensuring stability, interpretability, and robustness throughout the workflow.

For example, when collaborating with the \texttt{Explore-Agent}, the \texttt{PCS-Agent} operationalizes PCS principles by generating multiple perturbed versions of the dataset $\{D_1, D_2, \dots, D_k\}$ using different strategies (e.g., alternative imputation methods, outlier treatments, and feature transformations). Each perturbed dataset undergoes unit testing to verify semantic and structural validity. The system trains a corresponding model $\mathcal{M}_i$, forming a set of predictive fits (the pairing of an algorithm and a particular cleaned/preprocessed training dataset used for training the algorithm~\citep{vdsbook}) $\mathcal{F}_i = (D_i, \mathcal{M}_i)$.
These fits are compared on the basis of generalization performance and PCS-guided diagnostics, allowing the system to identify and report the most robust and reliable models.

Detailed functional modules and implementation of the \texttt{PCS-Agent}, including hypothesis generation, stability analysis, and visualization-based evaluations, are summarized in Appendix~\ref{tab:pcs-functions}. These functions provide the operational backbone for the critical thinking process described above, enabling systematic guidance and evaluation at all stages of the workflow.



\subsection{Tool Infrastructure and Execution Flow}

To support stable, modular, and reproducible execution in different stages of the pipeline, \texttt{VDSAgents} is equipped with an extensible tool infrastructure $\mathcal{T}$. 
These tools are available to all agents and serve key roles such as code execution, logic validation, error handling, and perturbation control.

\subsubsection{PCS-Guided ML Function Library.}
The core of the tool infrastructure is a modular ML function library $\mathcal{T}_\mathrm{ML}$, which supports data pre-processing, feature engineering, and structured perturbation.
It is used by $\mathcal{A}_\mathrm{explore}$ for cleaning and exploratory analysis, by $\mathcal{A}_\mathrm{model}$ for feature construction and model fitting, and by $\mathcal{A}_\mathrm{PCS}$ to generate perturbations.

Each function is implemented in a standalone format with explicit parameter interfaces and operation semantics.
The LLM can call these functions through natural language prompts by referencing predefined descriptions injected into the system message.
This design ensures consistent behavior across different agents and promotes traceability and reproducibility in multi-agent execution.
For a complete list of functions and their descriptions, see Appendix~\ref{append:mltools}.

\subsubsection{Unit Testing}
The unit test is a validation mechanism designed to systematically examine datasets for structural integrity and logical correctness~\citep{turkishtechnology2024unit}. Its primary purpose is to ensure that the processed datasets remain logically consistent and free of errors introduced during data perturbation and pre-processing steps.

After data cleaning, \texttt{Explore-Agent} invokes a suite of unit tests $\mathcal{U} = \{u_1, u_2, \dots, u_m\}$ to verify the structural and statistical validity of the dataset.
These tests detect issues such as missing values, unprocessed data loss, or duplicates. Each test outputs a structured result $\langle \text{name}, \text{passed}, \text{message} \rangle$ to guide downstream execution or debugging.
This mechanism reinforces computability and stability in early processing.
See Appendix~\ref{append:unit-tests} for test details.

\subsubsection{Code Execution and Debugging}

To ensure reliable code execution, each agent-generated script is handled by a \texttt{code executor}.
If execution fails or unit tests are not passed, the system invokes a \texttt{debug tool} that identifies errors, generates repair suggestions, and returns the corrected code.
This self-healing mechanism supports up to $N_{\max}$ retries (default value $N_{\max}=3$).
Upon repeated failures, it triggers human intervention or rolls back to the original planning state.
This design ensures robustness and recoverability in complex workflows. A schematic of this mechanism is provided in the Appendix~\ref{append:debug-flow}.

\subsubsection{Image-to-Text Support}

To enhance visual reasoning during EDA, \texttt{VDSAgents} incorporates an image-to-text module $\mathcal{T}_{\mathrm{OCR}}$ that extracts structured textual descriptions from graphical outputs such as histograms, box plots, and heatmaps.

Given an image input $I$, the module returns a set of textual elements:
\[
\mathcal{T}_{\mathrm{OCR}}(I) \rightarrow \mathcal{V}_{\mathrm{text}} = \{v_1, v_2, \dots, v_n\}.
\]
These include titles, axis labels, statistical extremes, trends, and outliers. The resulting set $\mathcal{V}_{\mathrm{text}}$ is then used by downstream agents (\texttt{Explore-Agent}, \texttt{PCS-Agent}) to assist in logic validation, anomaly interpretation and stability assessment.

\subsubsection{System Extensibility and Modularity}

\texttt{VDSAgents} is built with modularity and extensibility in mind.
On the model side, it defines an abstract interface $\mathcal{M}_{\mathrm{LLM}}$ that supports interchangeable use of various LLMs (e.g., ChatGPT, Claude, DeepSeek), allowing seamless switching without altering core logic.

On the tool side, the system maintains a dynamic set of modules:
\[
\mathcal{T} = \{\mathcal{T}_{\mathrm{ML}}, \mathcal{T}_{\mathrm{OCR}}, \mathcal{T}_{\mathrm{unit}}, \dots\},
\]
each registered with standardized interfaces for plug-and-play integration. Researchers can customize pipelines by adding domain-specific tools, pre-processing functions, or validation tests.

This architecture enables flexible adaptation to diverse tasks, from general-purpose modeling to specialized workflows such as time series forecasting or biomedical analysis—making \texttt{VDSAgents} a customizable and portable foundation for automated data science systems.

\subsection{Agent Function Interface and Mapping}

Each agent in the \texttt{VDSAgents} is equipped with a set of structured functions $\mathcal{F}_{\mathcal{A}_i} = \{f_1, f_2, \dots, f_m\}$, enabling it to perform domain-specific reasoning, code generation, and intermediate decision-making.
These functions can be called using natural language prompts and operate within a unified context composed of system messages, task instructions, and memory states.

The behavior of any agent $\mathcal{A}_i$ can be formalized as a functional mapping:
\[
\mathcal{A}_i: \quad (\mathcal{S}_{\text{context}}, \mathcal{F}_{\mathcal{A}_i}) \longrightarrow \mathcal{R}_{\text{task}},
\]
where
\begin{itemize}
  \item $\mathcal{S}_{\text{context}}$ represents the accumulated upstream outputs and task state;
  \item $\mathcal{F}_{\mathcal{A}_i}$ is the agent's internal callable function set;
  \item $\mathcal{R}_{\text{task}}$ is the resulting code, outputs, or structured reasoning reports.
\end{itemize}

Each function is designed to be modular, interpretable, and robust in perturbation. Appendix~\ref{append:agent-functions} provides detailed function listings for key agents.

\section{Experiments and Evaluation} 

\subsection{Experimental Setup} 

\subsubsection{Dataset} 

To evaluate the stability, robustness, and predictive performance of the proposed \texttt{VDSAgents}, we carry out experiments on nine representative datasets. These datasets range from clean, preprocessed data to raw data, allowing us to assess generalizability.

Let $\mathcal{D} = \{D_1, D_2, \dots, D_9\}$ denote the dataset collection, categorized as follows:

\begin{itemize}
  \item \textbf{Clean datasets} ($\mathcal{D}_\mathrm{clean}$): This group includes \texttt{bank\_churn}, \texttt{titanic}, and \texttt{obesity\_risks}, sourced from \href{https://www.kaggle.com}{Kaggle}. These datasets are already partially processed, with low missingness and consistent logical structures.

  \item \textbf{Raw datasets} ($\mathcal{D}_\mathrm{raw}$): Including \texttt{adult}, \texttt{In-Vehicle\_Coupon\_Recommendation}, \texttt{parkinsons}, and \texttt{Seoul\_Bike\_Sharing\_Demand}, these are drawn from the \href{https://archive.ics.uci.edu/}{UCI Machine Learning Repository}.
  They feature minimal pre-processing and present more realistic challenges, such as missing data and noisy attributes.

  \item \textbf{High-dimensional complex datasets} ($\mathcal{D}_\mathrm{complex}$): Consisting of \texttt{ames\_houses} and \texttt{online\_shopping}, both from \href{https://vdsbook.com/}{vdsbook.com}, these datasets are used in real-world educational or applied settings and involve intricate combinations of continuous and categorical features.
\end{itemize}

These datasets enable a comprehensive evaluation of \texttt{VDSAgents}' capabilities across various scenarios, particularly in tasks such as identification of response variables, selection of the feature pipeline, model comparison, and evaluation of the stability based on perturbations.
The complete datasets details are provided in Appendix~\ref{append:datasets}.

\subsubsection{Evaluation Metrics}

To systematically evaluate the performance of \texttt{VDSAgents} across diverse tasks and perturbed scenarios, we adopt the evaluation protocol introduced by~\citet{Hong2024DataInterpreter}, incorporating additional considerations regarding task performance and completion quality.
Three core metrics are defined below:


\begin{itemize}
    \item \textbf{Valid Submission (VS)}:  
    Measures the proportion of attempts in which the system successfully generates a syntactically correct, executable, and evaluable pipeline:
    \[
    \mathrm{VS} = \frac{T_s}{T},
    \]
    where $T_s$ is the number of successful attempts, and $T$ is the total number of attempts.  
    For each experiment, we repeatedly run the system until 5 valid outputs are obtained ($T_s = 5$). 
    As model responses may fail or be invalid, the value of $T$ varies across runs. 
    Thus, $VS = \frac{5}{T}$, enabling indirect inference of total attempts from reported VS values.

    \item \textbf{Average Normalized Performance Score (ANPS)}:  
    We first compute the Normalized Performance Score (NPS) for each valid run, and then report the mean and standard deviation across all $N$ valid runs. This ``run-level-first'' design enhances metric robustness and enables meaningful statistical reporting.
    
    \[
    \mathrm{NPS} =
    \begin{cases}
    \dfrac{1}{4} \left( \mathrm{Accuracy} + \mathrm{F1} + \mathrm{Precision} + \mathrm{Recall} \right), & \text{for classification tasks,} \\[12pt]
    \dfrac{1}{3} \left( \dfrac{1}{1 + \mathrm{RMSE}} + \dfrac{1}{1 + \mathrm{MAE}} + R^2 \right), & \text{for regression tasks.}
    \end{cases}
    \]
    \[
    \mathrm{ANPS} = \frac{1}{N} \sum_{i=1}^{N} \mathrm{NPS}^{(i)}, \quad
    \mathrm{SD}_{\mathrm{ANPS}} = \sqrt{\frac{1}{N} \sum_{i=1}^{N} \left(\mathrm{NPS}^{(i)} - \mathrm{ANPS} \right)^2}
    \]
    This `run-level-first'' approach enhances metric robustness and enables meaningful statistical reporting.
    For classification tasks, ANPS values are always between 0 and 1. For regression tasks, ANPS may become negative if $R^2_{\text{avg}} < 0$, which can occur in high-noise settings where model performance falls below baseline.

  \item \textbf{Comprehensive Score (CS)}:  
  Combines execution robustness and modeling performance into a unified metric:
  \[
  \mathrm{CS} = 0.5 \times \mathrm{VS} + 0.5 \times \mathrm{ANPS}.
  \]
  Equal weights are assigned to validity and quality, making CS suitable for comparing systems under heterogeneous data science tasks.
\end{itemize}

\subsubsection{Baselines and Model Configurations}

To benchmark the performance of \texttt{VDSAgents}, we compare it with two representative multi-stage automated data science frameworks: \texttt{AutoKaggle}~\citep{Li2024AutoKaggle} and \texttt{DataInterpreter}~\citep{Sun2024LAMBDA}.
For each system and dataset, experiments are repeated multiple times until five successful runs with valid outputs are obtained, and the final reported performance is averaged over these five runs.

We test both systems under two widely used LLM backends:

\begin{itemize}
  \item \textbf{GPT-4o}: A state-of-the-art OpenAI model with strong reasoning and code generation capabilities.
  \item \textbf{DeepSeek-V3}: A competitive open-source model representing leading domestic performance in structured tasks.
\end{itemize}

For all EDA-related visual analysis, we employ \textbf{Qwen-VL-7B} as the unified image-to-text module $\mathcal{T}_{\mathrm{OCR}}$ in both systems. 
We fix the maximum self-repair steps at $N_{\max}=3$ and set the number of perturbations to $k=50$.

\subsection{Results and Analysis}

This section presents a comprehensive evaluation of \texttt{VDSAgents} compared to \texttt{AutoKaggle} and \texttt{DataInterpreter} on nine benchmark tasks, comprising six classification datasets and three regression datasets.
The comparison focuses on three core dimensions that we have defined: VS, ANPS, and CS. 
Table~\ref{tab:metric-comparison-appendix} summarizes the performance in nine system configurations, each using one of two LLM back-ends: DeepSeek-V3 and GPT-4o. 

\subsubsection{Execution Stability}

VS measures the proportion of trials where the system successfully produces a valid, executable, and predictive output.
As shown in Table~\ref{tab:metric-comparison-appendix}, \texttt{VDSAgents} consistently achieves higher execution stability compared to \texttt{AutoKaggle} and \texttt{DataInterpreter}.
Specifically, with \texttt{DeepSeek-V3} and \texttt{GPT-4o}, \texttt{VDSAgents} attains average VS scores of 0.894 and 0.950 respectively, clearly outperforming \texttt{AutoKaggle} (0.577 and 0.534) and \texttt{DataInterpreter} (0.676 and 0.672).

Notably, \texttt{VDSAgents} with \texttt{GPT-4o} achieves 100\% success in eight of nine tasks, demonstrating strong robustness and compatibility with advanced LLMs. In contrast, both baseline methods exhibit lower and more variable performance, especially in challenging regression scenarios such as \texttt{parkinsons} and \texttt{online\_shopping} datasets.

Overall, \texttt{VDSAgents} offers a significantly more reliable execution framework for diverse tasks and conditions.

\subsubsection{Predictive Effectiveness}

ANPS reflects average predictive quality conditional on successful execution, combining classification accuracy and regression effectiveness.
As shown in Table~\ref{tab:metric-comparison-appendix}, \texttt{VDSAgents} consistently achieves higher ANPS scores than both \texttt{AutoKaggle} and \texttt{DataInterpreter}.
With \texttt{GPT-4o}, \texttt{VDSAgents} achieves an average ANPS of 0.692, clearly outperforming \texttt{AutoKaggle} (0.497) and \texttt{DataInterpreter} (0.569). Similarly, with \texttt{DeepSeek-V3}, \texttt{VDSAgents} obtains 0.667, surpassing \texttt{AutoKaggle} (0.599) and \texttt{DataInterpreter} (0.588).

In classification tasks, the three systems show relatively similar performance, with the average ANPS generally ranging between 0.7 and 0.9.
However, \texttt{VDSAgents} maintains slightly more stable and higher performance overall, particularly evident in datasets with noise or high dimensionality (e.g., online shopping).

In regression tasks, performance divergence is more significant.
\texttt{VDSAgents} with \texttt{GPT-4o} excels substantially (Parkinson’s ANPS=0.947, Seoul Bike ANPS=0.237), indicating its superior capability to model continuous numerical outputs.
In contrast, both \texttt{AutoKaggle} and \texttt{DataInterpreter} frequently deliver poor results, with \texttt{AutoKaggle-GPT4o} even yielding negative scores (e.g., Ames Houses), revealing clear limitations in numerical modeling capability and pipeline robustness.

These trends underscore the advantage of \texttt{VDSAgents}, particularly in regression contexts involving continuous variables, noise, or high dimensionality, where baseline methods struggle significantly.

Further details including NPS variability and ANPS visualizations with error bars are provided in Appendix~\ref{append:full-results}.

\subsubsection{Overall Capability}

\begin{figure}[htbp]
\centering
\includegraphics[width=13.5cm, height=9cm]{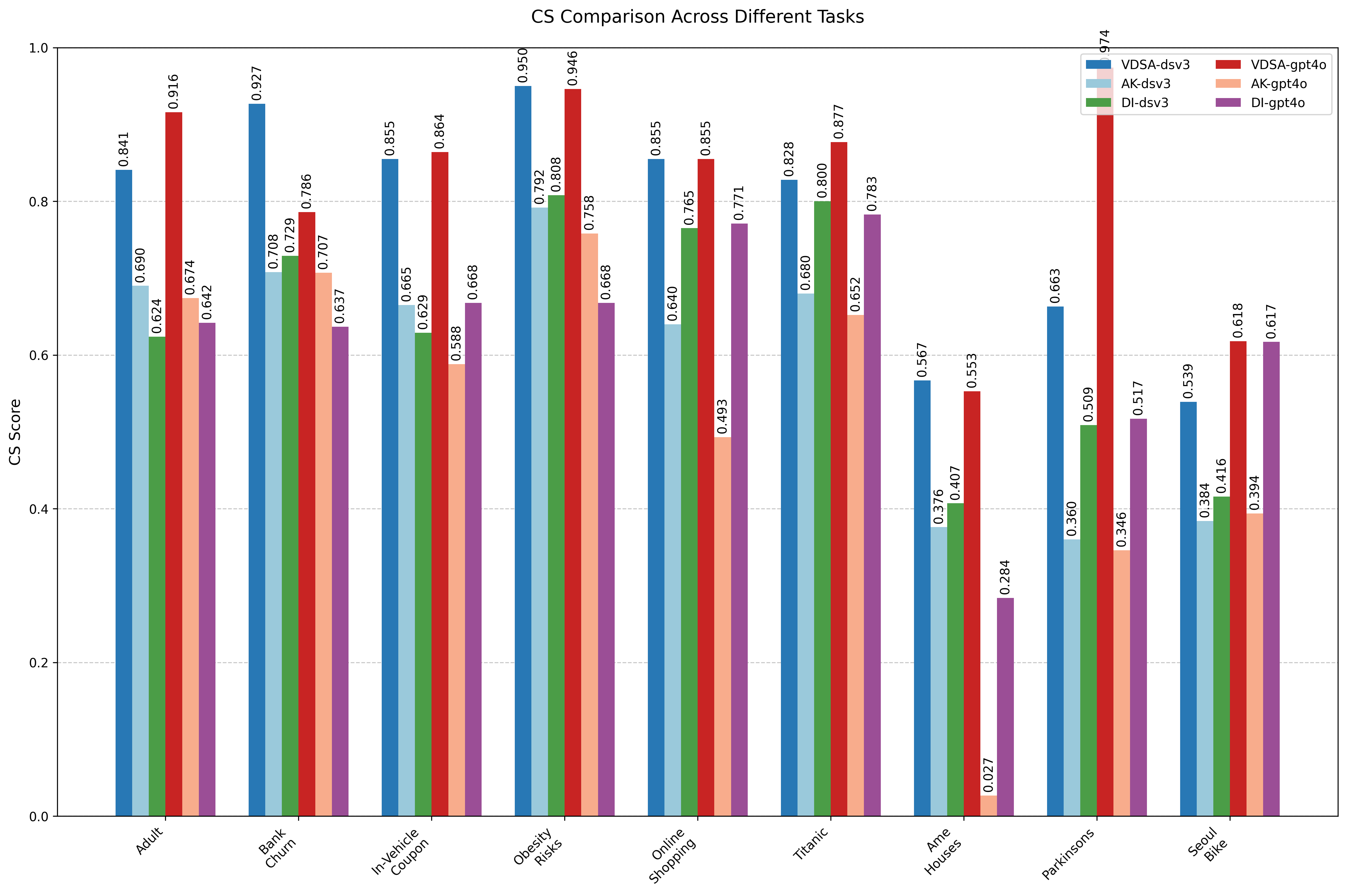}
\caption{Comparison of Comprehensive Score (CS) across four system variants on nine tasks}
\label{fig:cs-comparison}
\end{figure}

To jointly evaluate the stability and robustness of the execution and predictive effectiveness, we define CS as the average of VS and ANPS. 
As shown in Table~\ref{tab:metric-comparison-appendix} and Figure~\ref{fig:cs-comparison}, \texttt{VDSAgents} consistently achieves the highest CS scores compared to both \texttt{AutoKaggle} and \texttt{DataInterpreter}.
Specifically, \texttt{VDSAgents-GPT4o} attains an average CS of 0.821, clearly surpassing \texttt{AutoKaggle-GPT4o} (0.515) and \texttt{DataInterpreter-GPT4o} (0.621). Similar advantages persist with DeepSeek-V3.

In general, these results underscore the clear superiority of \texttt{VDSAgents}, offering robust execution combined with effective predictions in diverse tasks and challenging conditions.

\subsubsection{Ablation Study: Impact of PCS-Agent}

To further understand the contribution of key components within \texttt{VDSAgents}, we conducted an ablation study focusing on the role of the \texttt{PCS-Agent}—responsible for verifying Predictability, Computability, and Stability properties during pipeline generation.

We selected two representative datasets: Online Shopping (classification) and Ames Housing (regression).
We compared performance with and without the \texttt{PCS-Agent} module under the same evaluation protocol.

As shown in Table~\ref{tab:no_pcs_ablation}, removing the \texttt{PCS-Agent} results in substantial performance degradation, especially on the classification task (VS $\downarrow$ 20\%, ANPS $\downarrow$ 49\%, CS $\downarrow$ 31\%).
This demonstrates the importance of PCS auditing in maintaining robustness and predictive effectiveness across diverse tasks.

These results reinforce the critical role of PCS validation in improving both system reliability and predictive robustness.

For additional ablation results on key hyperparameters $k$ and $N_{max}$, please refer to Appendix~\ref{append:pcs-ablation}.

\section{Discussion}

The superior performance of \texttt{VDSAgents} over \texttt{AutoKaggle} and \texttt{DataInterpreter} is not only attributable to more powerful LLM back-ends, but to its principled design grounded in PCS principles. 
\texttt{VDSAgents} emphasizes three key elements of PCS: the ability to generate models that generalize (predictability), execute reliably (computability), and remain robust under perturbations (stability).
These principles are encoded in the logic of the \texttt{VDSAgents} at both the system and function levels.
Agents are not passive responders to prompts, but active planners that construct problem-solving trajectories aligned with the structure of the data and the goals of the analysis.

A concrete manifestation of this structure is observed in the way the system performs data cleaning.
For example, in missing value imputation, \texttt{AutoKaggle} typically applies global methods (e.g., filling with column-wise mean), overlooking latent data hierarchies.
This is a common pitfall in datasets where group structure matters, such as time series per stock or country-level survey data, where such methods can distort distributions and mislead downstream models.

In contrast, \texttt{VDSAgents}, under PCS-Guided, decomposes the cleaning process into reasoning steps: it first identifies the semantic roles of variables (e.g., \texttt{StockCode}, \texttt{Country}), then proposes targeted imputation strategies (e.g., per-stock mean, per-region median) that align with domain structure.
These strategies are not hallucinated, but supported by executable modular functions from the \texttt{mltools} library.

Furthermore, the \texttt{PCS-Agent} operationalizes stability by generating perturbed variants of the data based on alternative, yet plausible, structural assumptions, such as imputing by continent instead of country.
This enables the system to systematically assess the sensitivity of modeling outcomes, a core idea of the PCS principles that is often neglected in black-box pipelines.

By aligning each phase of the workflow, cleaning, modeling, evaluation, with a PCS-Guided structure, \texttt{VDSAgents} avoids the brittleness of purely prompt-driven systems.
It produces results that are not only accurate, but also reproducible, interpretable, and robust to variation.
In short, \texttt{VDSAgents} is not just ``LLM powered'' but ``PCS-Guided'', and this distinction is central to its observed advantages in diverse tasks and data types.
\section{Conclusion and Future Work}{}

This paper introduces \texttt{VDSAgents}, a modular, PCS-Guided, multi-agent automated data science framework. 
Guided by the core principles of predictability, computability, and stability, the system decomposes tasks into structured agent workflows, integrates reusable tools, and enables scientifically grounded modeling in real-world data scenarios.

Empirical results in nine datasets demonstrate the effectiveness of our design. \texttt{VDSAgents} achieves superior execution stability, robust predictive performance, and leading overall capability, outperforming the baseline \texttt{AutoKaggle} and \texttt{DataInterpreter} under both the GPT-4o and DeepSeek-V3 backends.
Its performance is especially notable on complex and noisy datasets, where its structured inference paths and stability-driven evaluation yield consistent gains.


Several promising directions remain open for extending the capabilities of \texttt{VDSAgents}:

\begin{itemize}
  \item \textbf{Fine-grained stability modeling:} Beyond basic data cleaning and feature selection, future work could explore stability-aware designs for more advanced modeling paths such as causal inference~\citep{wang2025epistasis} and multitask learning~\citep{agarwal2025pcs}.

  \item \textbf{Human-in-the-loop feedback:} Integrating expert feedback at key decision points could enable adaptive refinement of strategies and improve performance in domain-specific tasks.

  \item \textbf{Cross-domain generalization:} Applying the PCS-Guided architecture to critical domains such as healthcare, finance, or policy analysis will help evaluate its transferability and practical value under higher reliability demands.

\end{itemize}

By combining theoretical guidance with system-level engineering, \texttt{VDSAgents} offers a trustworthy foundation for LLM-driven data science.
We envision its broader applications in intelligent research, automated analysis, and education, helping bridge the gap between automation and scientific reasoning in data-driven practice.



\bibliography{bib/vds_agent}

\begin{thebibliography}{25}
\providecommand{\natexlab}[1]{#1}
\providecommand{\url}[1]{\texttt{#1}}
\expandafter\ifx\csname urlstyle\endcsname\relax
  \providecommand{\doi}[1]{doi: #1}\else
  \providecommand{\doi}{doi: \begingroup \urlstyle{rm}\Url}\fi

\bibitem[Achiam et~al.(2023)Achiam, Adler, Agarwal, Ahmad, Akkaya, Aleman, Almeida, Altenschmidt, Altman, Anadkat, et~al.]{achiam2023gpt}
Josh Achiam, Steven Adler, Sandhini Agarwal, Lama Ahmad, Ilge Akkaya, Florencia~Leoni Aleman, Diogo Almeida, Janko Altenschmidt, Sam Altman, Shyamal Anadkat, et~al.
\newblock {GPT-4} technical report.
\newblock \emph{arXiv preprint arXiv:2303.08774}, 2023.

\bibitem[Agarwal et~al.(2025)Agarwal, Xiao, Barter, Ronen, Fan, and Yu]{agarwal2025pcs}
Abhineet Agarwal, Michael Xiao, Rebecca Barter, Omer Ronen, Boyu Fan, and Bin Yu.
\newblock {PCS-UQ: Uncertainty Quantification via the Predictability-Computability-Stability Framework}.
\newblock \emph{arXiv preprint arXiv:2505.08784}, 2025.

\bibitem[Cai et~al.(2023)Cai, Bai, Zha, and Chen]{cai2023toolmaker}
Cheng Cai, Lu~Bai, Shengyu Zha, and Enhong Chen.
\newblock {LLMs as Tool Makers: Teaching LLMs to Create and Use Tools}.
\newblock \emph{arXiv preprint arXiv:2312.06644}, 2023.

\bibitem[Cao(2017)]{Cao2017}
Longbing Cao.
\newblock Data science: a comprehensive overview.
\newblock \emph{ACM Computing Surveys (CSUR)}, 50\penalty0 (3):\penalty0 1--42, 2017.
\newblock \doi{10.1145/3076253}.

\bibitem[DeepSeek-AI(2024)]{deepseek-llm}
DeepSeek-AI.
\newblock {DeepSeek LLM: Scaling Open-Source Language Models with Longtermism}.
\newblock \emph{arXiv preprint arXiv:2401.02954}, 2024.

\bibitem[Gao et~al.(2023)Gao, Madaan, Zhou, Alon, Liu, Yang, Callan, and Neubig]{Gao2022PAL}
Luyu Gao, Aman Madaan, Shuyan Zhou, Uri Alon, Pengfei Liu, Yiming Yang, Jamie Callan, and Graham Neubig.
\newblock {PAL: Program-aided Language Models}.
\newblock In \emph{Proceedings of the 40th International Conference on Machine Learning}, volume 202, pages 10764--10799, 2023.

\bibitem[Hong et~al.(2024)Hong, Lin, Liu, Liu, Wu, Zhang, Wei, Li, Chen, Zhang, Wang, Zhang, Zhang, Yang, Zhuge, Guo, Zhou, Tao, Tang, Lu, Zheng, Liang, Fei, Cheng, Gou, Xu, and Wu]{Hong2024DataInterpreter}
Sirui Hong, Yizhang Lin, Bang Liu, Bangbang Liu, Binhao Wu, Ceyao Zhang, Chenxing Wei, Danyang Li, Jiaqi Chen, Jiayi Zhang, Jinlin Wang, Li~Zhang, Lingyao Zhang, Min Yang, Mingchen Zhuge, Taicheng Guo, Tuo Zhou, Wei Tao, Xiangru Tang, Xiangtao Lu, Xiawu Zheng, Xinbing Liang, Yaying Fei, Yuheng Cheng, Zhibin Gou, Zongze Xu, and Chenglin Wu.
\newblock {Data Interpreter: An LLM Agent For Data Science}.
\newblock \emph{arXiv preprint arXiv:2402.18679}, 2024.

\bibitem[Jiang et~al.(2025)Jiang, Schmidt, Srikanth, Xu, Kaplan, Jacenko, and Wu]{aide2025}
Zhengyao Jiang, Dominik Schmidt, Dhruv Srikanth, Dixing Xu, Ian Kaplan, Deniss Jacenko, and Yuxiang Wu.
\newblock {AIDE: AI-Driven Exploration in the Space of Code}.
\newblock \emph{arXiv preprint arXiv:2502.13138}, 2025.

\bibitem[Li et~al.(2024)Li, Zang, Ma, Guo, Zheng, Liu, Niu, Wang, Yang, Liu, Zhong, Zhou, Huang, and Zhang]{Li2024AutoKaggle}
Ziming Li, Qianbo Zang, David Ma, Jiawei Guo, Tuney Zheng, Minghao Liu, Xinyao Niu, Yue Wang, Jian Yang, Jiaheng Liu, Wanjun Zhong, Wangchunshu Zhou, Wenhao Huang, and Ge~Zhang.
\newblock {AutoKaggle: A Multi-Agent Framework for Autonomous Data Science Competitions}.
\newblock \emph{arXiv preprint arXiv:2410.20424}, 2024.

\bibitem[Madaan et~al.(2023)Madaan, Tandon, Gupta, Hallinan, Gao, Wiegreffe, Alon, Dziri, Prabhumoye, Yang, Gupta, Majumder, Hermann, Welleck, Yazdanbakhsh, and Clark]{Madaan2023SelfRefine}
Aman Madaan, Niket Tandon, Prakhar Gupta, Skyler Hallinan, Luyu Gao, Sarah Wiegreffe, Uri Alon, Nouha Dziri, Shrimai Prabhumoye, Yiming Yang, Shashank Gupta, Bodhisattwa~Prasad Majumder, Katherine Hermann, Sean Welleck, Amir Yazdanbakhsh, and Peter Clark.
\newblock {Self-Refine: Iterative Refinement with Self-Feedback}.
\newblock In \emph{Advances in Neural Information Processing Systems 36 (NeurIPS 2023)}, volume~36, pages 46534--46594, 2023.

\bibitem[Provost and Fawcett(2013)]{provost2013science}
Foster Provost and Tom Fawcett.
\newblock Data science and its relationship to big data and data-driven decision making.
\newblock \emph{Big Data}, 1\penalty0 (1):\penalty0 51--59, 2013.
\newblock \doi{10.1089/big.2013.1508}.

\bibitem[Qian et~al.(2023)Qian, Chen, Jiang, Yang, Zhang, Song, Tan, and Zhang]{qian2023creator}
Yuxuan Qian, Shuchang Chen, Yichong Jiang, Zeqi Yang, Jiani Zhang, Yangqiu Song, Hao Tan, and Xiaodan Zhang.
\newblock {Creator: A Unified Agent Framework for Tool Creation and Usage}.
\newblock \emph{arXiv preprint arXiv:2310.01753}, 2023.

\bibitem[Schick et~al.(2023)Schick, Dwivedi-Yu, Sun, Webson, Hou, Chaffin, Behnke, Lewis, Sch{\"a}rli, Scales, et~al.]{schick2023toolformer}
Timo Schick, Jack Dwivedi-Yu, Leandro von~Werra Sun, Albert Webson, Yujie Hou, Alexander~M Chaffin, Sven Behnke, Patrick Lewis, Nathanael Sch{\"a}rli, Nathan Scales, et~al.
\newblock {Toolformer: Language Models Can Teach Themselves to Use Tools}.
\newblock \emph{arXiv preprint arXiv:2302.04761}, 2023.

\bibitem[Sun et~al.(2024)Sun, Han, Jiang, Qi, Sun, Yuan, and Huang]{Sun2024LAMBDA}
Maojun Sun, Ruijian Han, Binyan Jiang, Houduo Qi, Defeng Sun, Yancheng Yuan, and Jian Huang.
\newblock {LAMBDA: A Large Model Based Data Agent}.
\newblock \emph{arXiv preprint arXiv:2407.17535}, 2024.

\bibitem[Technology(2024)]{turkishtechnology2024unit}
Turkish Technology.
\newblock {Integrating Unit Testing into Data Science Lifecycle}.
\newblock \url{https://medium.com/@turkishtechnology/integrating-unit-testing-into-data-science-lifecycle-6de01a33a4c0}, December 2024.
\newblock Medium, published December 12, 2024. Accessed July 25, 2025.

\bibitem[Tu et~al.(2024)Tu, Zou, Su, and Zhang]{Tu2024LLMDataScience}
Xinming Tu, James Zou, Weijie~J. Su, and Linjun Zhang.
\newblock {What Should Data Science Education Do with Large Language Models?}
\newblock \emph{Harvard Data Science Review}, 6\penalty0 (1), 2024.
\newblock \doi{10.1162/99608f92.bff007ab}.

\bibitem[Wang et~al.(2023)Wang, Lin, Liu, Liu, and Liu]{wang2023selfdebug}
Bohan Wang, Xiang Lin, Peng Liu, Zhiwei~Steven Liu, and Yang Liu.
\newblock {Code-as-Policies: Self-Debugging LLMs with Unit Tests}.
\newblock \emph{arXiv preprint arXiv:2305.13242}, 2023.

\bibitem[Wang et~al.(2025)Wang, Tang, Youlton, Weldy, Kenney, Ronen, Hughes, Chin, Sutton, Agarwal, et~al.]{wang2025epistasis}
Qianru Wang, Tiffany~M Tang, Michelle Youlton, Chad~S Weldy, Ana~M Kenney, Omer Ronen, J~Weston Hughes, Elizabeth~T Chin, Shirley~C Sutton, Abhineet Agarwal, et~al.
\newblock {Epistasis regulates genetic control of cardiac hypertrophy}.
\newblock \emph{Nature Cardiovascular Research}, pages 1--21, 2025.

\bibitem[Yao et~al.(2023)Yao, Yu, Zhao, Yu, and Narasimhan]{yao2023tree}
Shinn Yao, Dian Yu, Jeffrey Zhao, Bill Yu, and Karthik Narasimhan.
\newblock {Tree of Thoughts: Deliberate Problem Solving with Large Language Models}.
\newblock \emph{arXiv preprint arXiv:2305.10601}, 2023.

\bibitem[Yao et~al.(2022)Yao, Zhao, Yu, Du, Shafran, Narasimhan, and Cao]{yao2022react}
Shunyu Yao, Jeffrey Zhao, Dian Yu, Nan Du, Izhak Shafran, Karthik Narasimhan, and Yuan Cao.
\newblock {ReAct: Synergizing Reasoning and Acting in Language Models}.
\newblock \emph{arXiv preprint arXiv:2210.03629}, 2022.

\bibitem[Yu(2013)]{Yu2013Stability}
Bin Yu.
\newblock Stability.
\newblock \emph{Bernoulli}, 19\penalty0 (4):\penalty0 1484--1500, 2013.
\newblock \doi{10.3150/13-BEJSP14}.

\bibitem[Yu and Barter(2024)]{vdsbook}
Bin Yu and Rebecca~L. Barter.
\newblock {Veridical Data Science: The Practice of Responsible Data Analysis and Decision Making}.
\newblock \url{https://vdsbook.com/}, 2024.
\newblock [Online; accessed 2025-05-15].

\bibitem[Yu and Kumbier(2020)]{Yu2020VDS}
Bin Yu and Karl Kumbier.
\newblock Veridical data science.
\newblock \emph{Proceedings of the National Academy of Sciences}, 117\penalty0 (8):\penalty0 3920--3929, 2020.
\newblock \doi{10.1073/pnas.1901326117}.

\bibitem[Zhou et~al.(2023{\natexlab{a}})Zhou, Zhang, Xu, Yang, Lin, Pan, and Liu]{zhou2023autoagents}
Yi~Zhou, Yichi Zhang, Yi~Xu, Zhi Yang, Zheng Lin, Liang Pan, and Zhiyuan Liu.
\newblock {AutoAgents: Self-improving LLM Agents via Uncertainty and Self-Reflection}.
\newblock \emph{arXiv preprint arXiv:2309.00661}, 2023{\natexlab{a}}.

\bibitem[Zhou et~al.(2023{\natexlab{b}})Zhou, Tang, Yuan, Xu, Liu, Tang, and Zhou]{zhou2023whiteboxtest}
Ziqi Zhou, Yunyu Tang, Zehao Yuan, Canwen Xu, Zhiyuan Liu, Jialiang Tang, and Jie Zhou.
\newblock {White-box Testing for LLMs via Execution Verification}.
\newblock \emph{arXiv preprint arXiv:2307.03868}, 2023{\natexlab{b}}.

\end{thebibliography}
\bibliographystyle{plainnat}

\appendix
\begin{appendix}
    \onecolumn
    \begin{center}
        {\huge Supplementary Material to ``VDSAgents: A PCS-Guided Multi-Agent System for Veridical Data Science Automation''}
    \end{center}

    \section{Prompt Templates for \texttt{PCS-Agent}}
    \label{append:prompt-templates}

    This appendix presents the structured prompt templates used in the \texttt{PCS-Agent}, designed to embed the reasoning logic of PCS into agent-level planning and evaluation.

    \subsection{System Message Template}
    \label{append:system-message}

    \begin{tcolorbox}[title={System Message for \texttt{PCS-Agent}}, 
        colback=gray!5, colframe=black, sharp corners=south, 
        fonttitle=\bfseries, title filled=true]

    \textbf{Role:} You are a data science expert responsible for evaluating other agents' outputs based on the PCS-Guided framework and issuing critical feedback.

    \vspace{1ex}
    \textbf{1. Key Definitions (Excerpt):}
    \begin{itemize}
        \item \textbf{PCS-Guided Framework:} A principle-based framework for evaluating Predictability, Computability, and Stability across the data science lifecycle.
        \item \textbf{Predictability:} Whether conclusions generalize to new or external data.
        \item \textbf{Stability:} Sensitivity of conclusions to changes in data or methodology.
        \item \textbf{Computability:} Practical feasibility of executing analytical steps.
    \end{itemize}

    \vspace{1ex}
    \textbf{2. Evaluation Suggestions (Excerpt):}
    \begin{itemize}
        \item For EDA: consider validating findings with external data or literature.
        \item For modeling: create perturbed versions and compare predictive fits.
    \end{itemize}

    \vspace{1ex}
    \textbf{3. Context Information:}
    \begin{itemize}
        \item \texttt{Problem description: \{problem\_description\}}
        \item \texttt{Data context: \{context\_description\}}
    \end{itemize}

    \end{tcolorbox}

    \subsection{Task-Specific Message Template}
    \label{append:task-message}

    \begin{tcolorbox}[title={Task Message for PCS-Guided Evaluation (e.g., EDA)}, 
        colback=gray!5, colframe=black, sharp corners=south, 
        fonttitle=\bfseries, title filled=true]

    \textbf{Task:} Analyze the PCS-Guided properties of EDA conclusions.

    \textbf{Description:} Based on execution results, evaluate the Predictability, Stability, and Computability of the EDA outputs, and return a structured assessment report.

    \vspace{1ex}
    \textbf{Input (excerpt):}
    \begin{itemize}
        \item \texttt{Conclusion: \{conclusion\}}
        \item \texttt{Evaluation Results: \{result\}}
    \end{itemize}

    \vspace{1ex}
    \textbf{Expected Output:}
    \begin{lstlisting}[language=json]
    [
      {
        "Predictability": "Assessment of generalizability to unseen data",
        "Stability": "Evaluation under input/data perturbation"
      }
    ]
    \end{lstlisting}

    \end{tcolorbox}

\section{Function Tables for \texttt{VDSAgents} Components}
\label{append:tool-tables}

This appendix collects the reference tables of the core utility functions used in the \texttt{VDSAgents} system. These include modules for machine learning preprocessing, data perturbation, and validation. Each table summarizes the functions' names, their descriptions, and the scope of usage between different agents. The tools are designed for modularity, extensibility, and alignment with the PCS-Guided framework.

\subsection{ML Function Library}
\label{append:mltools}
Table~\ref{tab:ml_tools_summary} lists key functions in the \texttt{mltools} module used for cleaning, transforming, and engineering features during the \texttt{explore} and \texttt{model} stages.

\begin{table}[H]
\centering
\caption{Overview of ML Function Library (\texttt{mltools})}
\label{tab:ml_tools_summary} 
\begin{tabularx}{\textwidth}{lX}
\toprule
\textbf{Function Name} & \textbf{Description} \\
\midrule
\texttt{fill\_missing} & Impute missing values using mean, median, KNN, or group-wise logic \\
\texttt{handle\_outliers} & Detect and correct outliers via IQR, Z-score, or quantile filtering \\
\texttt{encode\_categorical} & Convert categorical variables using label, one-hot, or frequency encoding \\
\texttt{remove\_columns} & Drop features based on missingness, low variance, or correlation \\
\texttt{transform\_features} & Apply transformations (log, min-max, standard scaling) \\
\texttt{discretize\_features} & Bin continuous variables using equal-width, quantiles, or KMeans \\
\texttt{select\_features} & Perform feature selection (e.g., mutual info, variance, Lasso, RFE) \\
\texttt{create\_polynomial\_features} & Generate higher-order or interaction features \\
\texttt{reduce\_dimensions} & Reduce dimensionality via PCA or LDA \\
\bottomrule
\end{tabularx}
\end{table}

\subsection{Unit Test Functions}
\label{append:unit-tests}
Table~\ref{tab:unit-tests} lists unit tests used to validate data quality and structure after cleaning. These tests are invoked by the \texttt{explore} agent to ensure that outputs meet basic consistency requirements.

\begin{table}[H]
\centering
\caption{Unit Tests for Cleaned Dataset Validation}
\label{tab:unit-tests}
\begin{tabularx}{\textwidth}{lX}
\toprule
\textbf{Test Name} & \textbf{Description} \\
\midrule
\texttt{test\_file\_readable} & Check whether output file exists and is readable \\
\texttt{test\_empty\_dataset} & Detect empty datasets with column headers but no rows \\
\texttt{test\_missing\_values} & Identify unprocessed missing values and report proportions \\
\texttt{test\_duplicated\_features} & Detect duplicated column names \\
\texttt{test\_duplicated\_rows} & Detect identical duplicate samples \\
\texttt{test\_data\_consistency} & Compare schema before and after cleaning \\
\texttt{test\_data\_retention} & Ensure sufficient row retention rate (default: >85\%) \\
\bottomrule
\end{tabularx}
\end{table}

\subsection{Function Lists for Core Agents}
\label{append:agent-functions}

The following tables summarize the core functions embedded in the five core agents. 
Each function supports parameterized usage and natural language triggering.

\subsubsection{\texttt{Define-Agent}}
The \texttt{Define-Agent} focuses on clarifying the problem context, 
loading preliminary data, and evaluating the relevance of the variable to guide downstream modeling tasks. 
Its core functions are summarized in Table~\ref{tab:define-functions}.

\begin{table}[H]
\centering
\caption{Function Modules for \texttt{Define-Agent}}
\label{tab:define-functions}
\begin{tabularx}{\textwidth}{lX}
\toprule
\textbf{Function Name} & \textbf{Description} \\
\midrule
\texttt{load\_data\_preview} & Load first rows and extract variable names \\
\texttt{analyze\_variables} & Analyze variables and generate descriptions \\
\texttt{detect\_observation\_unit} & Detect dataset observation unit \\
\texttt{evaluate\_variable\_relevance} & Evaluate relevance of variables \\
\texttt{execute\_problem\_definition} & Run full problem definition pipeline \\
\bottomrule
\end{tabularx}
\end{table}

\subsubsection{\texttt{Explore-Agent}}
The \texttt{Explore-Agent} performs data cleaning and exploratory data analysis (EDA), 
covering invalid/missing value detection, cleaning operations, and exploratory questions. 
Its core functions are summarized in Table~\ref{tab:explore-functions}.

\begin{table}[H]
\centering
\caption{Function Modules for \texttt{Explore-Agent}}
\label{tab:explore-functions}
\begin{tabularx}{\textwidth}{lX}
\toprule
\textbf{Function Name} & \textbf{Description} \\
\midrule
\texttt{generate\_cleaning\_task\_list} & Create task list for cleaning workflow \\
\texttt{generate\_dimension\_check\_code} & Code for checking dataset dimensions \\
\texttt{analyze\_data\_dimension} & Analyze dimension check results \\
\texttt{check\_for\_invalid\_values} & Detect invalid values in dataset \\
\texttt{generate\_missing\_value\_analysis\_code} & Code for missing value analysis \\
\texttt{analyze\_missing\_values\_result} & Analyze missing value results \\
\texttt{generate\_data\_integrity\_check\_code} & Code for data integrity check \\
\texttt{analyze\_and\_generate\_fillna\_operations} & Generate cleaning ops from integrity check \\
\texttt{generate\_cleaning\_operations} & Merge problem list into cleaning ops \\
\texttt{generate\_hypothesis\_validation\_code} & Code for hypothesis validation \\
\texttt{analyze\_hypothesis\_validation\_result} & Analyze hypothesis validation results \\
\texttt{generate\_cleaning\_code} & Generate complete cleaning code \\
\texttt{execute\_cleaning\_operations} & Run cleaning operations \\
\texttt{generate\_eda\_questions} & Formulate EDA questions \\
\texttt{generate\_eda\_code} & Code for EDA questions \\
\texttt{analyze\_eda\_result} & Analyze EDA results \\
\texttt{solve\_eda\_questions} & Solve EDA questions end-to-end \\
\texttt{generate\_pcs\_evaluation\_code} & Code for PCS evaluation \\
\texttt{check\_discrete\_variables} & Check if discrete variables need encoding \\
\texttt{generate\_discrete\_variable\_code} & Code for encoding discrete variables \\
\texttt{load\_and\_compare\_data} & Compare samples to validate data \\
\texttt{execute\_cleaning\_tasks} & Execute full cleaning task sequence \\
\texttt{analyze\_image} & Analyze visualization images \\
\texttt{generate\_eda\_summary} & Generate EDA summary report \\
\bottomrule
\end{tabularx}
\end{table}

\subsubsection{\texttt{Model-Agent}}
The \texttt{Model-Agent} is responsible for proposing modeling strategies 
and generating executable pipelines based on the input dataset and exploratory analysis results. 
Its core functions are summarized in Table~\ref{tab:model-functions}.

\begin{table}[H]
\centering
\caption{Function Modules for \texttt{Model-Agent}}
\label{tab:model-functions}
\begin{tabularx}{\textwidth}{lX}
\toprule
\textbf{Function Name} & \textbf{Description} \\
\midrule
\texttt{identify\_response\_variable} & Detect response variable and its type \\
\texttt{suggest\_feature\_engineering\_methods} & Recommend feature engineering strategies \\
\texttt{suggest\_modeling\_methods} & Suggest ranked modeling approaches \\
\texttt{generate\_combined\_model\_code} & Generate code combining models and features \\
\texttt{train\_and\_evaluate\_combined\_models} & Train and evaluate multiple models \\
\texttt{execute\_batch\_evaluation} & Run evaluation across perturbed datasets \\
\texttt{summarize\_evaluation\_results} & Summarize model performance \\
\bottomrule
\end{tabularx}
\end{table}

\subsubsection{\texttt{Evaluate-Agent}}
The \texttt{Evaluate-Agent} is designed to assess model stability and performance, 
generate best-fit datasets, and produce evaluation codes to validate model results. 
Its core functions are summarized in Table~\ref{tab:evaluate-functions}.

\begin{table}[H]
\centering
\caption{Function Modules for \texttt{Evaluate-Agent}}
\label{tab:evaluate-functions}
\begin{tabularx}{\textwidth}{lX}
\toprule
\textbf{Function Name} & \textbf{Description} \\
\midrule
\texttt{generate\_test\_datasets\_code} & Code to create best-fit datasets \\
\texttt{generate\_and\_execute\_test\_datasets} & Generate and run best-fit datasets workflow \\
\texttt{generate\_model\_evaluation\_code} & Code for model training and evaluation \\
\texttt{generate\_and\_execute\_model\_evaluation} & Generate and run model evaluation workflow \\
\bottomrule
\end{tabularx}
\end{table}

\subsubsection{\texttt{PCS-Agent}}
The \texttt{PCS-Agent} performs PCS evaluations, 
including hypothesis generation, stability analysis, and visualization interpretation. 
Its core functions are summarized in Table~\ref{tab:pcs-functions}.

\begin{table}[H]
\centering
\caption{Function Modules for \texttt{PCS-Agent}}
\label{tab:pcs-functions}
\begin{tabularx}{\textwidth}{lX}
\toprule
\textbf{Function Name} & \textbf{Description} \\
\midrule
\texttt{analyze\_image} & Analyze visualization images \\
\texttt{analyze\_pcs\_evaluation\_result} & Evaluate conclusions using PCS principles \\
\texttt{evaluate\_problem\_definition} & Assess problem definition and generate hypotheses \\
\texttt{generate\_stability\_analysis\_code} & Code for data cleaning stability analysis \\
\texttt{execute\_stability\_analysis} & Run stability analysis and validate datasets \\
\bottomrule
\end{tabularx}
\end{table}

\section{Debugging Mechanism}
\label{append:debug-flow}

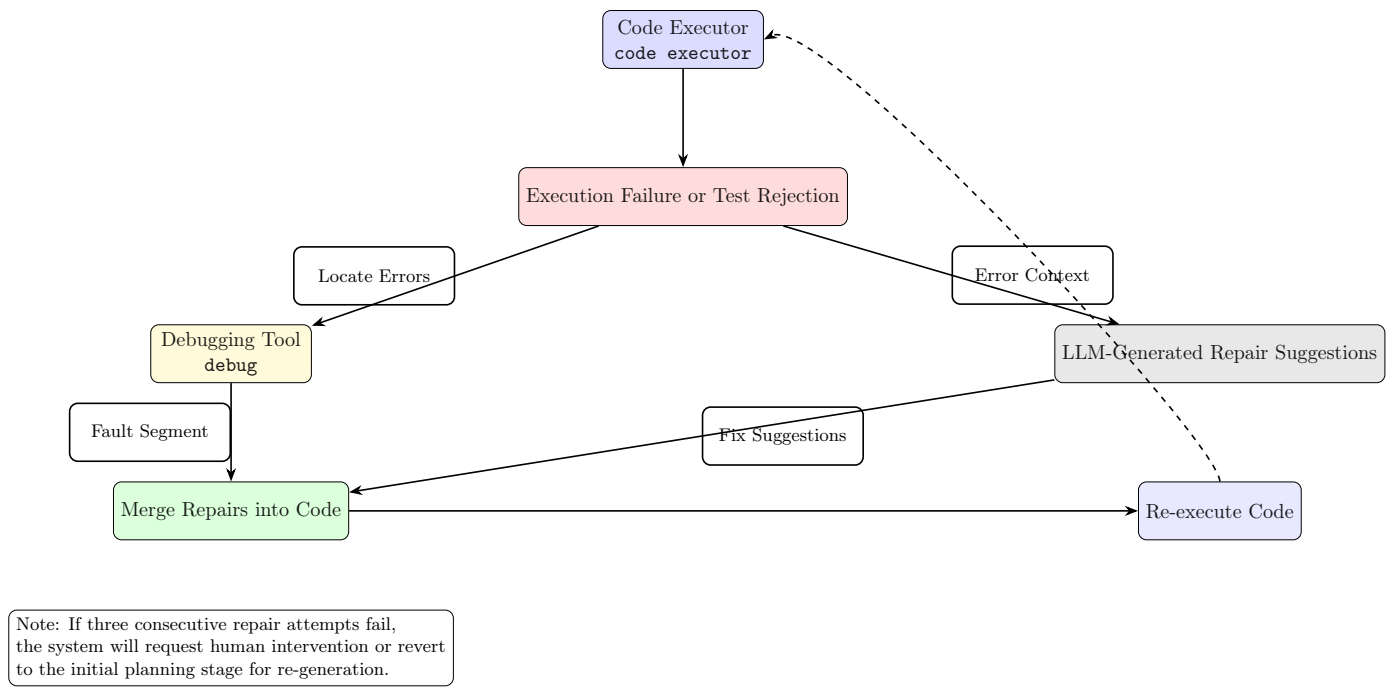
\begin{figure}[htbp]
\centering
\resizebox{18.5cm}{9cm}{  
\begin{tikzpicture}[
  node distance=1.7cm and 2.6cm,
  every node/.style={
    font=\small, 
    rounded corners, 
    draw, 
    minimum width=2.8cm, 
    minimum height=1.0cm, 
    align=center
  },
  >=Stealth
  ]

  \node[fill=blue!15, fill opacity=0.9] (executor) {Code Executor\\ \texttt{code executor}};
  \node[fill=red!15, fill opacity=0.9, below=of executor] (error) {Execution Failure or Test Rejection};
  \node[fill=yellow!20, fill opacity=0.9, below left=of error, xshift=-1cm] (debugtool) {Debugging Tool\\ \texttt{debug}};
  \node[fill=gray!20, fill opacity=0.9, below right=of error, xshift=1cm] (llm) {LLM-Generated Repair Suggestions};
  \node[fill=green!15, fill opacity=0.9, below=of debugtool] (merge) {Merge Repairs into Code};
  \node[fill=blue!10, fill opacity=0.9, below=of llm] (reexec) {Re-execute Code};

  \draw[->, thick] (executor) -- (error);
  \draw[->, thick] (error) -- (debugtool) node[midway, left] {\footnotesize Locate Errors};
  \draw[->, thick] (error) -- (llm) node[midway, right] {\footnotesize Error Context};
  \draw[->, thick] (debugtool) -- (merge) node[midway, left] {\footnotesize Fault Segment};
  \draw[->, thick] (llm) -- (merge) node[midway, right] {\footnotesize Fix Suggestions};
  \draw[->, thick] (merge) -- (reexec);
  \draw[->, thick, dashed] (reexec.north) .. controls +(0,0.7) and +(1,0.5) .. (executor.east);

  \node[below=1.2cm of merge, font=\footnotesize, align=left] (note) {
    Note: If three consecutive repair attempts fail,\\
    the system will request human intervention or revert\\
    to the initial planning stage for re-generation.
  };

\end{tikzpicture}
}
\caption{Debugging workflow between the code executor and LLM-based repair module.}
\label{fig:debug-flow-appendix}
\end{figure}

Figure~\ref{fig:debug-flow-appendix} illustrates the collaborative mechanism between the code executor and the debugging tool in the \texttt{VDSAgents} system. When execution fails or unit tests are not passed, the system invokes an automated repair loop, guided by structured error messages and repair suggestions from the LLM.

\section{Datasets}
\label{append:datasets}

This appendix summarizes the datasets used in our experiments, including source links, prediction targets, and feature descriptions.

\begin{enumerate}
  \item \textbf{Adult Income Prediction}  
  Source: \url{https://archive.ics.uci.edu/dataset/2/adult}  
  Based on the 1994 U.S. Census, this dataset aims to predict whether an individual's annual income exceeds \$50,000. The target variable \texttt{income} is binary (\texttt{"<=50K"} or \texttt{">50K"}). Evaluation metric: Precision. Features include age, education, occupation, race, gender, capital gain/loss, weekly working hours, and native country (15 variables in total).

  \item \textbf{Bank Customer Churn}  
  Source: \url{https://www.kaggle.com/competitions/playground-series-s4e1}  
  This classification task predicts whether a bank customer will churn (\texttt{Exited = 1}). The evaluation metric is AUC. Features include credit score, country, gender, age, tenure, balance, number of products, credit card status, activity status, and estimated salary.

  \item \textbf{In-Vehicle Coupon Recommendation}  
  Source: \url{https://archive.ics.uci.edu/dataset/603/in+vehicle+coupon+recommendation}  
  Collected via Amazon Mechanical Turk, this dataset simulates a driving scenario to predict whether a driver will accept a coupon (\texttt{Y = 1}). Evaluation metric: Accuracy. It contains 26 contextual and behavioral features such as destination, passengers, weather, time, gender, marital status, education, occupation, income, and entertainment frequency.

  \item \textbf{Obesity Risk Prediction}  
  Source: \url{https://www.kaggle.com/competitions/playground-series-s4e2}  
  This dataset aims to classify individuals into obesity risk categories (\texttt{NObeyesdad}) based on dietary habits, exercise frequency, and basic physiological indicators. Evaluation metric: Accuracy. Features include BMI, food consumption frequency, alcohol intake, and sedentary time.

  \item \textbf{Online Shopping Intentions}  
  Source: \url{https://vdsbook.com/}  
  This binary classification task predicts whether a browsing session will result in a purchase. The target variable indicates whether a transaction occurred. Evaluation metric: Accuracy. Features include counts of page visits (administrative, informational, product), time spent per type, returning visitor flag, browser type, and weekend indicator (17 features in total).

  \item \textbf{Titanic Survival Prediction}  
  Source: \url{https://www.kaggle.com/competitions/titanic}  
  Based on the 1912 Titanic disaster, this dataset predicts whether a passenger survived (\texttt{Survived}). Evaluation metric: Accuracy. Features include gender, age, passenger class, fare, embarkation port, and family relationships.

  \item \textbf{Ames Housing Prices}  
  Source: \url{https://vdsbook.com/}  
  A regression task to predict house sale prices (\texttt{SalePrice}) in Ames, Iowa. Evaluation metric: RMSE. Features include house area, construction year, garage, basement, remodeling quality, and other structured attributes—suitable for explainable modeling.

  \item \textbf{Parkinson’s Telemonitoring}  
  Source: \url{https://archive.ics.uci.edu/dataset/189/parkinsons+telemonitoring}  
  This dataset includes voice-based biomedical features from 42 patients to predict two continuous scores: \texttt{motor\_UPDRS} and \texttt{total\_UPDRS}. Evaluation metric: Mean Squared Error (MSE). Total samples: 5,875. Features include fundamental frequency, amplitude variations, spectral complexity, and tremor intensity.

  \item \textbf{Seoul Bike Sharing Demand}  
  Source: \url{https://archive.ics.uci.edu/dataset/560/seoul+bike+sharing+demand}  
  This dataset records the hourly rental counts in Seoul (2017-2018), along with the weather and calendar features. The target variable is \texttt{Rented Bike Count}. Suitable for time-series regression. Evaluation metric: $R^2$. The features include temperature, humidity, wind speed, solar radiation, rainfall, season, and holiday indicators (14 variables).
\end{enumerate}

\section{Full Experimental Results}
\label{append:full-results}

Table~\ref{tab:metric-comparison-appendix} presents a detailed comparison of three automated systems, \texttt{VDSAgents}, \texttt{AutoKaggle}, and \texttt{DataInterpreter}, in nine datasets using three evaluation metrics: Valid Submission Rate (VS), Average Normalized Performance Score (ANPS), and Comprehensive Score (CS). Results are reported separately for classification and regression tasks, under two model configurations: deepseekv3 and gpt-4o. For brevity, we use abbreviations VDSA, AK, and DI to denote \texttt{VDSAgents}, \texttt{AutoKaggle}, and \texttt{DataInterpreter}, respectively, in the table.

\begin{table}[htbp]
\centering
\small
\caption{Comparison of VS, ANPS, and CS across nine datasets for different systems}
\label{tab:metric-comparison-appendix}
\renewcommand{\arraystretch}{1.2}
\setlength{\tabcolsep}{4pt}
\begin{threeparttable}
\begin{tabular}{ll|ccccccccc|c}
\toprule[1.5pt]
\textbf{Metric} & \textbf{System} & \multicolumn{6}{c}{Classification Tasks} & \multicolumn{3}{c}{Regression Tasks} & Avg. \\
\cmidrule(lr){3-8} \cmidrule(lr){9-11}
& & Ad & BC & IVC & OR & OS & Tit & Hou & Park & SeB & \\
\midrule
\multirow{6}{*}{VS}
& VDSA-dsv3     & 0.833 & \textbf{1.000} & \textbf{1.000} & \textbf{1.000} & \textbf{1.000} & 0.833 & \textbf{0.833} & 0.714 & 0.833 & 0.894 \\
& AK-dsv3       & 0.556 & 0.556 & 0.625 & 0.714 & 0.625 & 0.556 & 0.455 & 0.556 & 0.556 & 0.577 \\
& DI-dsv3 & 0.556 & 0.625 & 0.556 & 0.714 & 0.833 & 0.833 & 0.625 & 0.625 & 0.714 & 0.676 \\
& VDSA-gpt4o    & \textbf{1.000} & 0.714 & \textbf{1.000} & \textbf{1.000} & \textbf{1.000} & \textbf{1.000} & \textbf{0.833} & \textbf{1.000} & \textbf{1.000} & \textbf{0.950} \\
& AK-gpt4o      & 0.556 & 0.556 & 0.500 & 0.625 & 0.500 & 0.556 & 0.455 & 0.455 & 0.556 & 0.534 \\
& DI-gpt4o & 0.455 & 0.556 & 0.625 & 0.556 & 0.833 & 0.833 & 0.357 & 0.833 & 1.000 & 0.672 \\
\midrule
\multirow{6}{*}{ANPS}
& VDSA-dsv3     & \textbf{0.848} & 0.855 & 0.709 & 0.900 & 0.709 & \textbf{0.823} & \textbf{0.301} & 0.611 & \textbf{0.245} & 0.667 \\
& AK-dsv3       & 0.824 & \textbf{0.861} & 0.706 & 0.870 & 0.655 & 0.804 & 0.298 & 0.165 & 0.213 & 0.599 \\
& DI-dsv3 & 0.692 & 0.834 & 0.703 & \textbf{0.902} & 0.698 & 0.766 & 0.189 & 0.392 & 0.117 & 0.588 \\
& VDSA-gpt4o    & 0.832 & 0.857 & \textbf{0.728} & 0.891 & \textbf{0.710} & 0.753 & 0.273 & \textbf{0.947} & 0.237 & \textbf{0.692} \\
& AK-gpt4o      & 0.792 & 0.859 & 0.677 & 0.890 & 0.485 & 0.748 & -0.447 & 0.237 & 0.232 & 0.497 \\
& DI-gpt4o & 0.829 & 0.718 & 0.710 & 0.779 & 0.709 & 0.732 & 0.210 & 0.201 & 0.234 & 0.569 \\
\midrule
\multirow{6}{*}{CS}
& VDSA-dsv3     & 0.841 & \textbf{0.927} & 0.855 & \textbf{0.950} & \textbf{0.855} & 0.828 & \textbf{0.567} & 0.663 & 0.539 & 0.780 \\
& AK-dsv3       & 0.690 & 0.708 & 0.665 & 0.792 & 0.640 & 0.680 & 0.376 & 0.360 & 0.384 & 0.588 \\
& DI-dsv3 & 0.624 & 0.729 & 0.629 & 0.808 & 0.765 & 0.800 & 0.407 & 0.509 & 0.416 & 0.632 \\
& VDSA-gpt4o    & \textbf{0.916} & 0.786 & \textbf{0.864} & 0.946 & \textbf{0.855} & \textbf{0.877} & 0.553 & \textbf{0.974} & \textbf{0.618} & \textbf{0.821} \\
& AK-gpt4o      & 0.674 & 0.707 & 0.588 & 0.758 & 0.493 & 0.652 & 0.027 & 0.346 & 0.394 & 0.515 \\
& DI-gpt4o & 0.642 & 0.637 & 0.668 & 0.668 & 0.771 & 0.783 & 0.284 & 0.517 & 0.617 & 0.621 \\
\bottomrule[1.5pt]
\end{tabular}
\begin{tablenotes}
\small
\item \textbf{Notes:} Ad = Adult, BC = Bank Churn, IVC = In-Vehicle Coupon, OR = Obesity Risks, OS = Online Shopping, Tit = Titanic, Hou = Ames Houses, Park = Parkinson's, SeB = Seoul Bike. Bold entries indicate best performance for each metric-task combination.
\end{tablenotes}
\end{threeparttable}
\end{table}

Table~\ref{tab:no_pcs_ablation} presents the results of an ablation study conducted to assess the contribution of the \texttt{PCS-Agent} module within the \texttt{VDSAgents} system.

\begin{table}[h]
\centering
\caption{Ablation results: with vs. without PCS-Agent.}
\label{tab:no_pcs_ablation}
\begin{tabular}{l|ccc|ccc}
\toprule
\multirow{2}{*}{Dataset} & \multicolumn{3}{c|}{Without PCS-Agent} & \multicolumn{3}{c}{With PCS-Agent} \\
 & VS & ANPS & CS & VS & ANPS & CS \\
\midrule
Online Shopping & 0.8333 & 0.4748 & 0.6541 & \textbf{1.0000} & \textbf{0.7090} & \textbf{0.8550} \\
Ames Housing    & 0.8333 & 0.2486 & 0.5410 & \textbf{0.8333} & \textbf{0.3010} & \textbf{0.5670} \\
\bottomrule
\end{tabular}
\end{table}

Table~\ref{tab:anps_cv_all} reports the coefficient of variation (CV) of the NPS for 5 valid runs for all datasets and systems. 
CV is calculated as the ratio of the standard deviation to the mean of NPS, and serves as a normalized measure of variability. 
We report these values under both the DeepSeek-V3 and GPT-4o model configurations for \texttt{VDSAgent}, \texttt{AutoKaggle}, and \texttt{DataInterpreter}.

This analysis complements our robustness evaluation by quantifying the stability of predictive performance between multiple valid executions. 
A lower CV indicates more consistent performance across runs, while a higher CV reveals potential volatility in the results, especially under different model configurations or datasets. 
The results show that \texttt{VDSAgent} tends to exhibit lower variability in most settings compared to other systems, reinforcing the effectiveness of PCS-guided validation and design.

\begin{table}[h]
\centering
\caption{NPS coefficient of variation (CV) across 5 valid runs for all datasets and systems under DeepSeek-V3 and GPT-4o.}
\label{tab:anps_cv_all}
\small
\begin{tabular}{lcccccc}
\toprule
\multirow{2}{*}{\textbf{Dataset}} & \multicolumn{2}{c}{\texttt{VDSAgent}} & \multicolumn{2}{c}{\texttt{AutoKaggle}} & \multicolumn{2}{c}{\texttt{DataInterpreter}} \\
\cmidrule(lr){2-3} \cmidrule(lr){4-5} \cmidrule(lr){6-7}
& DeepSeek-V3 & GPT-4o & DeepSeek-V3 & GPT-4o & DeepSeek-V3 & GPT-4o \\
\midrule
Adult               & 0.0242 & 0.0519 & 0.0081 & 0.0716 & 0.5001 & 0.0415 \\
Bank Churn          & 0.0046 & 0.0055 & 0.0019 & 0.0033 & 0.0157 & 0.0054 \\
In-Vehicle Coupon   & 0.0213 & 0.0225 & 0.0580 & 0.0904 & 0.1250 & 0.0895 \\
Obesity Risks       & 0.0037 & 0.0108 & 0.0213 & 0.0071 & 0.0007 & 0.2119 \\
Online Shopping     & 0.0175 & 0.0096 & 0.1317 & 0.3746 & 0.0092 & 0.0075 \\
Titanic             & 0.0227 & 0.0247 & 0.0714 & 0.0391 & 0.1591 & 0.0757 \\
AmeHouses           & 0.0060 & 0.0909 & 0.0272 & -3.2387 & 1.1533 & 0.4502 \\
Parkinsons          & 0.3868 & 0.0485 & 1.6558 & 4.0418 & 0.8484 & 6.0423 \\
Seoul Bike          & 0.0253 & 0.0651 & 0.1449 & 0.0406 & 2.3826 & 0.1733 \\
\bottomrule
\end{tabular}
\end{table}

\begin{figure}
\centering
\includegraphics[width=0.9\textwidth]{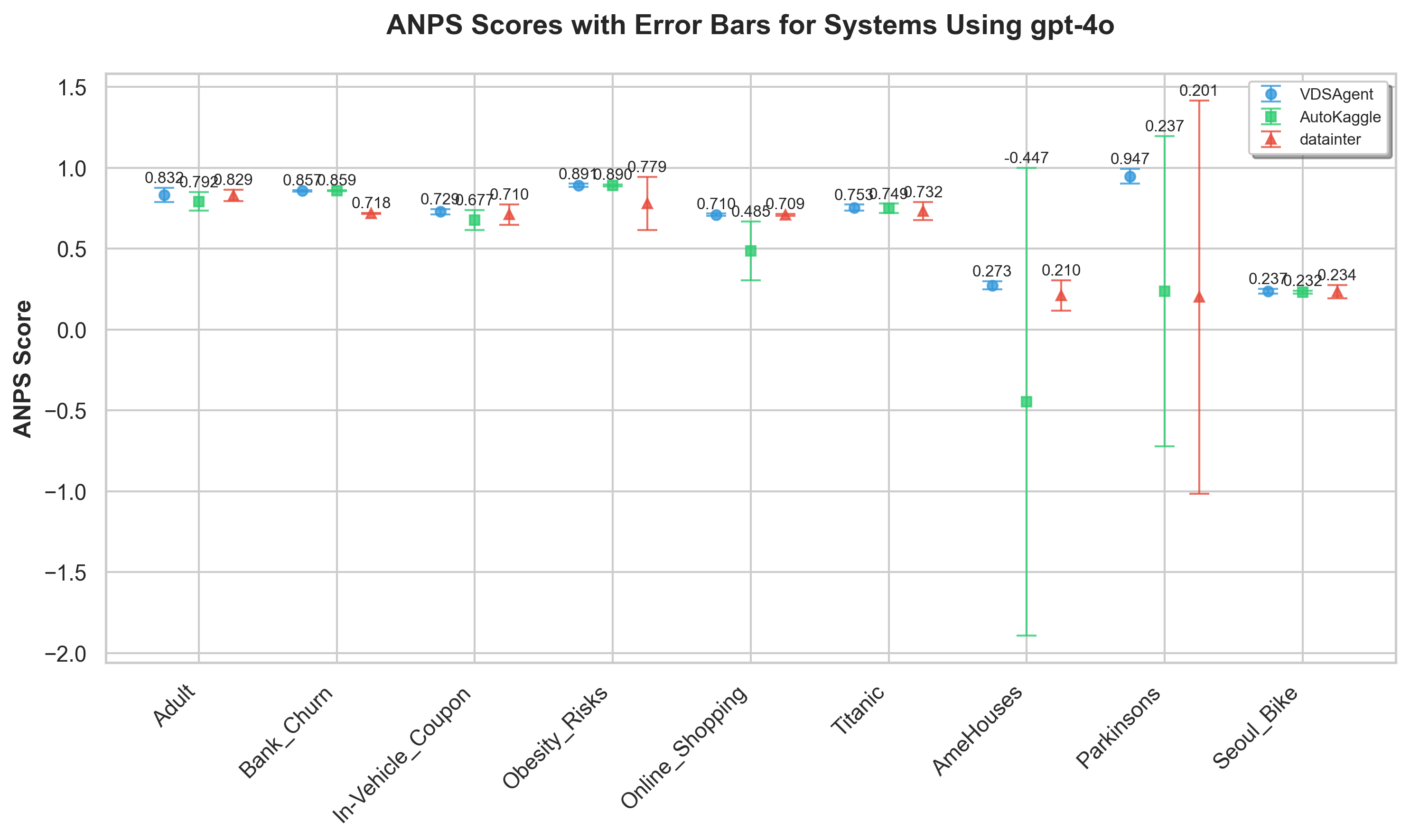}
\caption{ANPS comparison under GPT-4o backend with standard deviation error bars.}
\label{fig:anps_gpt4o}
\end{figure}

\begin{figure}
\centering
\includegraphics[width=0.9\textwidth]{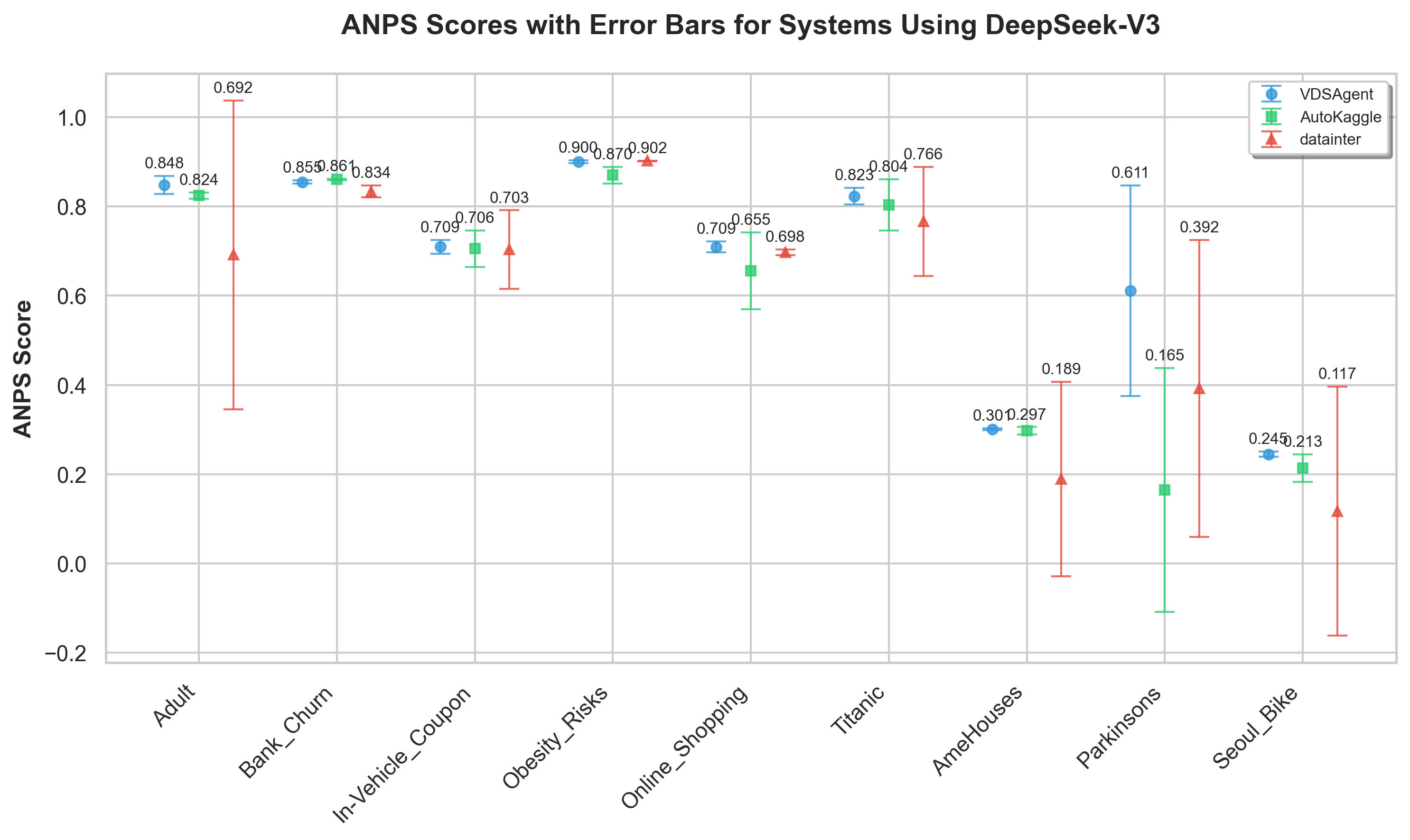}
\caption{ANPS comparison under DeepSeek-V3 backend with standard deviation error bars.}
\label{fig:anps_deepseek}
\end{figure}

Figures~\ref{fig:anps_gpt4o} and \ref{fig:anps_deepseek} visualize ANPS performance across datasets under different LLM backends.
Each bar includes standard deviation error bars computed from 5 valid runs, which quantify the variability in performance and highlight the robustness of each system on each task.

\section{Ablation Study on PCS Parameters}
\label{append:pcs-ablation}

To further understand the contribution of PCS-based mechanisms, we carried out ablation experiments on two core hyperparameters: the perturbation count $k$ and the maximum number of self-repair steps $N_{max}$.
Experiments were run on two representative datasets: Online Shopping (classification) and Ames Housing (regression).

\subsection{Effect of Perturbation Count $k$}

We evaluated the performance of ANPS at different values of $k \in \{3, 4, 5, 8, 10\}$. As shown in Table~\ref{tab:k_ablation} and the left panel of Figure~\ref{fig:k_n_ablation}, increasing $k$ generally improves performance, with gains saturating around $k{=}5$.
This indicates that moderate diversity in perturbations strengthens PCS auditing, but excessive perturbations provide limited additional value.

\begin{table}[h]
\centering
\caption{ANPS (mean ± std) across different values of $k$ for two datasets.}
\label{tab:k_ablation}
\begin{tabular}{c|cc|cc}
\toprule
\multirow{2}{*}{$k$} & \multicolumn{2}{c|}{Online Shopping} & \multicolumn{2}{c}{Ames Housing} \\
 & ANPS & Std & ANPS & Std \\
\midrule
3  & 0.5709 & 0.1052 & 0.1825 & 0.2277 \\
4  & 0.5759 & 0.1186 & 0.1819 & 0.2274 \\
5  & 0.5890 & 0.0993 & 0.2891 & 0.0097 \\
8  & 0.6015 & 0.1331 & 0.2897 & 0.0036 \\
10 & 0.7069 & 0.0110 & 0.2970 & 0.0047 \\
\bottomrule
\end{tabular}
\end{table}

\subsection{Effect of Maximum Self-Repair Steps $N_{max}$}

\begin{figure}
\centering
\includegraphics[width=0.9\linewidth]{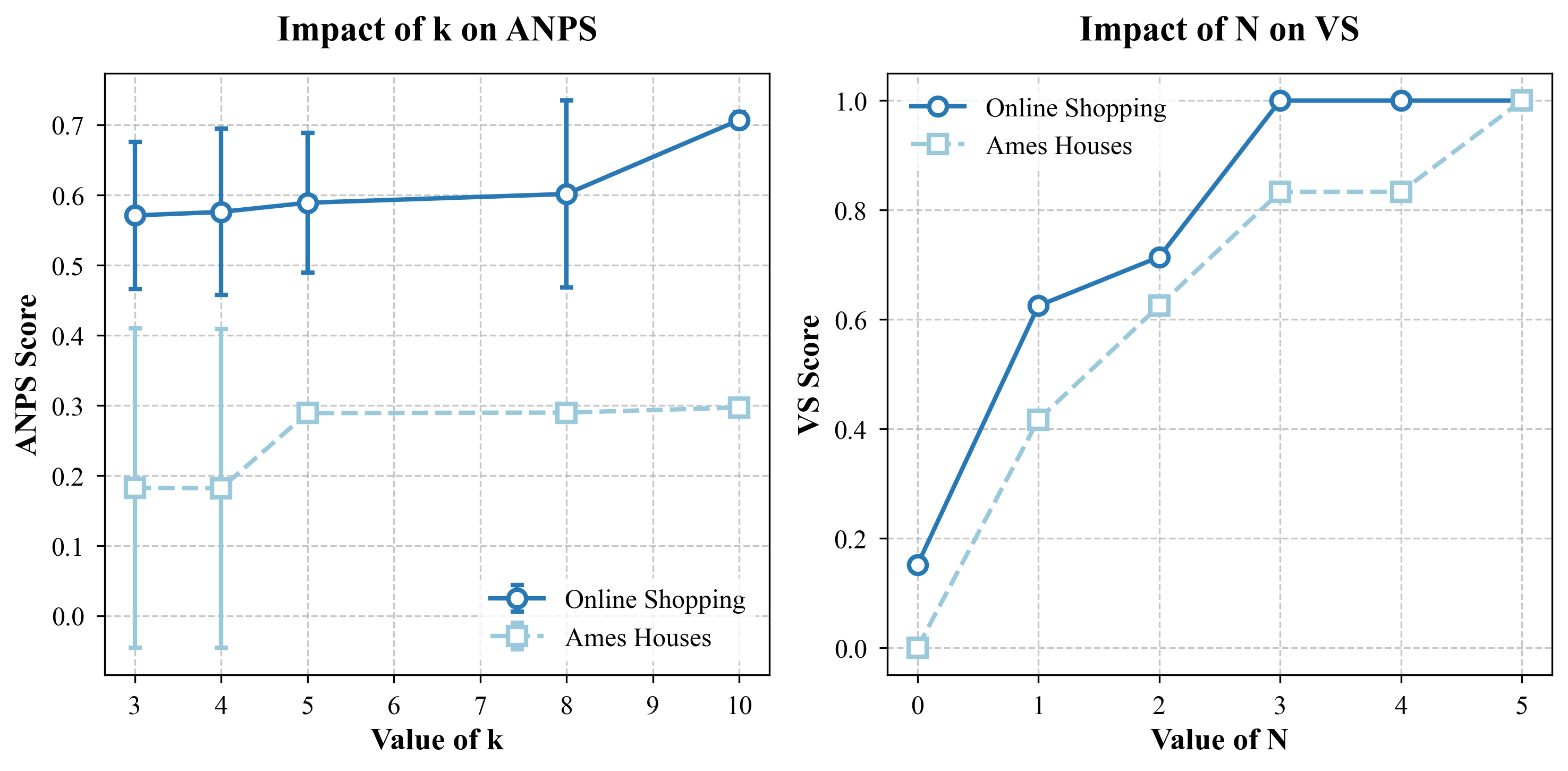}
\caption{(Left) ANPS vs. $k$ on two datasets; (Right) VS vs. $N_{max}$. Error bars show standard deviation across 5 runs.}
\label{fig:k_n_ablation}
\end{figure}

We then investigated the impact of $N_{max}$ by evaluating VS in values from 0 to 5. Table~\ref{tab:nmax_ablation} and the right panel of Figure~\ref{fig:k_n_ablation} show that even a small number of self-repair steps can substantially improve the valid submission rate, especially in challenging scenarios. This validates the role of self-repair in improving \emph{completability and robustness}, even if the prediction quality (ANPS) remains relatively stable.

\begin{table}[h]
\centering
\caption{VS scores across different values of $N_{max}$ for two datasets.}
\label{tab:nmax_ablation}
\begin{tabular}{c|c|c}
\toprule
$N_{max}$ & VS (Online Shopping) & VS (Ames Housing) \\
\midrule
0 & 0.1515 & 0.0000 \\
1 & 0.6250 & 0.4167 \\
2 & 0.7143 & 0.6250 \\
3 & 1.0000 & 0.8333 \\
4 & 1.0000 & 0.8333 \\
5 & 1.0000 & 1.0000 \\
\bottomrule
\end{tabular}
\end{table}

Together, these ablations validate the effectiveness of both PCS perturbation auditing ($k$) and self-repair mechanisms ($N_{max}$), contributing to overall system robustness and execution success.

\end{appendix}

\end{document}